\documentclass[10pt,twocolumn,letterpaper]{article}

\usepackage{cvpr}              %
\usepackage[accsupp]{axessibility}
\usepackage{graphicx}
\usepackage{amsmath}
\usepackage{amssymb}
\usepackage{booktabs}

\usepackage[pagebackref,breaklinks,colorlinks]{hyperref}

\usepackage[capitalize]{cleveref}
\crefname{section}{Sec.}{Secs.}
\Crefname{section}{Section}{Sections}
\Crefname{table}{Table}{Tables}
\crefname{table}{Tab.}{Tabs.}

\usepackage{url}
\usepackage{subcaption}
\usepackage{amsmath,amsthm,amssymb}
\usepackage{xcolor}
\usepackage{autobreak}

\usepackage{wrapfig}
\usepackage{multirow}
\usepackage{multicol}
\usepackage{microtype}
\usepackage{fixltx2e}
\usepackage{comment}
\usepackage{graphicx}
\usepackage{bm}
\usepackage{booktabs} %

\usepackage{amsmath}
\usepackage{amssymb}
\usepackage{algorithm}
\usepackage{algpseudocode}
\usepackage{amsthm}
\usepackage{nameref}

\usepackage[american]{babel}

\usepackage{color}

\usepackage{graphicx}
\allowdisplaybreaks

\def\eqref#1{equation~\ref{#1}}

\def\onedot{.}
\def\eg{\emph{e.g}\onedot} 

\def\ie{\emph{i.e}\onedot}

\def\etal{\emph{et al}\onedot}

\DeclareMathOperator*{\argmin}{argmin}
\DeclareMathOperator*{\argmax}{argmax}

\newtheorem{theorem}{Theorem}

\newtheorem{definition}[theorem]{Definition}
\newtheorem{lemma}[theorem]{Lemma}

\newtheorem{claim}[theorem]{Claim}

\newtheorem{remark}{Remark}

\newcounter{mylabelcounter}

\makeatletter
\newcommand{\labelText}[2]{%
#1\refstepcounter{mylabelcounter}%
\immediate\write\@auxout{%
  \string\newlabel{#2}{{1}{\thepage}{{\unexpanded{#1}}}{mylabelcounter.\number\value{mylabelcounter}}{}}%
}%
}
\makeatother

\newcommand{\w}{\bm{w}}
\newcommand{\x}{\bm{x}}

\renewcommand{\comment}[1]{}

\newcommand{\cN}{\mathcal{N}}

\newcommand{\bbE}{\mathbb{E}}

\newcommand{\bbI}{\mathbb{I}}

\newcommand{\erf}{\textup{\textrm{erf}}}

\newlength\myindent %
\algdef{SE}[SUBALG]{Indent}{EndIndent}{}{\algorithmicend\ }%
\algtext*{Indent}
\algtext*{EndIndent}

\newcommand{\ym}[1]{\textcolor{blue}{(ym: #1)}}

\def\w{{\mathbf{w}}}
\def\x{{\mathbf{x}}}

\usepackage{graphicx}

\begin{document}

\title{Cooperation or Competition: Avoiding Player Domination for Multi-Target Robustness via Adaptive Budgets} %
\author{
Yimu Wang\\
University of Waterloo\\
Waterloo, Canada\\
{\tt\small yimu.wang@uwaterloo.ca}
\and
Dinghuai Zhang\\
Mila, University of Montreal\\
Montreal, Canada\\
{\tt\small dinghuai.zhang@mila.quebec}
\and
Yihan Wu\\
University of Pittsburgh\\
Pittsburgh, United States\\
{\tt\small yiw154@pitt.edu}
\and
Heng Huang\\
University of Pittsburgh\\
Pittsburgh, United States\\
{\tt\small henghuanghh@gmail.com}
\and
Hongyang Zhang \thanks{Corresponding author.}\\
University of Waterloo\\
Waterloo, Canada\\
{\tt\small hongyang.zhang@uwaterloo.ca}
} %
\maketitle

\begin{abstract}
    Despite incredible advances, deep learning has been shown to be susceptible to adversarial attacks. Numerous approaches have been proposed to train robust networks both empirically and certifiably. However, most of them defend against only a single type of attack, while recent work takes steps forward in defending against multiple attacks. 
    In this paper, to understand multi-target robustness, we view this problem as a bargaining game in which different players (adversaries) negotiate to reach an agreement on a joint direction of parameter updating. 
    We identify a phenomenon named \emph{player domination} in the bargaining game, namely that the existing max-based approaches, such as MAX and MSD, do not converge. Based on our theoretical analysis, we design a novel framework that adjusts the budgets of different adversaries to avoid any player dominance. 
    Experiments on standard benchmarks show that employing the proposed framework to the existing approaches significantly advances multi-target robustness.
\end{abstract}

\section{Introduction}

\begin{figure}[t]
    \begin{center}
    \includegraphics[width=0.4\textwidth]{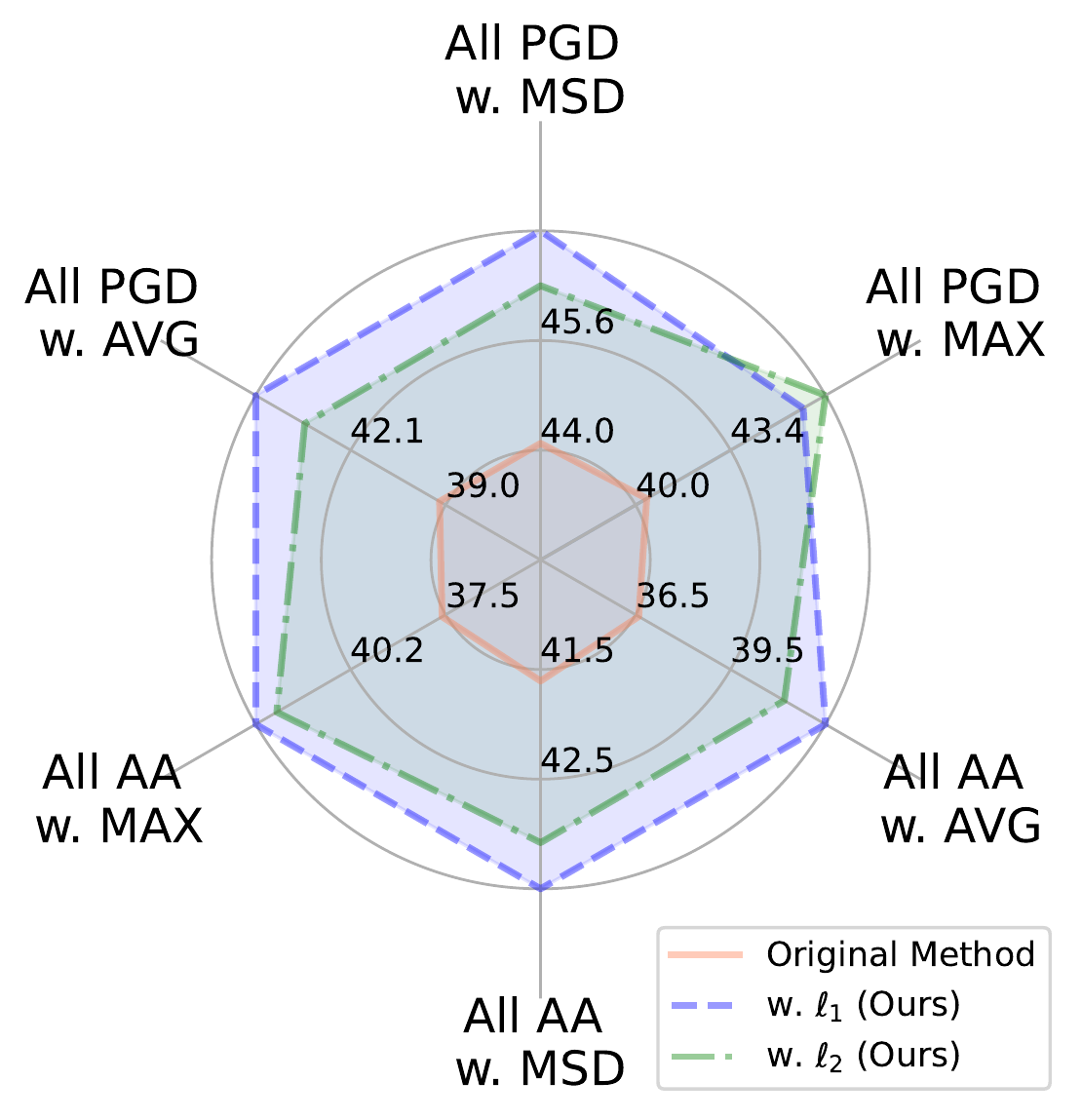}
    \vspace{-1.3em}
        \caption{Robust accuracy against PGD attacks and AutoAttack (``AA'' in this figure)  on CIFAR-10. ``All'' means that the model successfully defends against the $\ell_1$, $\ell_2$, and $\ell_\infty$ (PGD or AutoAttack) attacks simultaneously. Compared with the previously best-known methods, our proposed framework achieves improved performance. ``w. $\ell_1$'' and ``w. $\ell_2$'' refer to the model training with our proposed \textbf{AdaptiveBudget} algorithm with $\ell_1$ or $\ell_2$ norms, respectively.}\label{fig:intro}
    \end{center}
    \vspace{-2.3em}
\end{figure}

Machine learning (ML) models~\cite{DBLP:conf/cvpr/HeZRS16,10.1145/3394171.3413882,9093306} have been shown to be susceptible to adversarial examples~\cite{szegedyIntriguingPropertiesNeural2014}, where human-imperceptible perturbations added to a clean example might arbitrarily change the output of machine learning models. Adversarial examples are generated by maximizing the loss within a small perturbation region around a clean example, \eg, $\ell_\infty$, $\ell_1$ and $\ell_2$ balls. On the other hand, numerous heuristic defenses have been proposed to be robust against adversarial examples, \eg, distillation~\cite{papernotDistillationDefenseAdversarial2016}, logit-pairing~\cite{kannanAdversarialLogitPairing2018} and adversarial training~\cite{madryDeepLearningModels2018}.

However, most of the existing defenses are only robust against one type of attacks~\cite{madryDeepLearningModels2018,raghunathanCertifiedDefensesAdversarial2022,wongProvableDefensesAdversarial2018,engstromRotationTranslationSuffice2018}, while they fail to defend against other adversaries. For example, existing work~\cite{kangTransferAdversarialRobustness2019,mainiAdversarialRobustnessUnion2020} showed that robustness in the $\ell_p$ threat model does not necessarily generalize to other $\ell_q$ threat models when $p \neq q$. However, for the sake of the safety of ML systems, it has been argued that one should target robustness against multiple adversaries simultaneously~\cite{croceProvableRobustnessAll2020}.

Recently, various methods~\cite{schott2018towards,tramerAdversarialTrainingRobustness2019a,mainiAdversarialRobustnessUnion2020} have been proposed to address this problem. Multi-target adversarial training, which targets defending against multiple adversarial perturbations, has attracted significant attention: 
a variational autoencoder-based model~\cite{schott2018towards} learns a classifier robust to multiple perturbations; after that, MAX and AVG strategies, which aggregate different adversaries for adversarial training against multiple threat models, have been shown to enjoy improved performance~\cite{tramerAdversarialTrainingRobustness2019a}. To further advance the robustness against multiple adversaries, MSD~\cite{mainiAdversarialRobustnessUnion2020} is proposed and outperformed MAX and AVG by taking the worst case over all steepest descent directions. 
These methods follow a general scheme similar to the (single-target) adversarial training. They first sample adversarial examples by different adversaries and then update the model with the aggregation of the gradients from these adversarial examples.

This general scheme for multi-target adversarial training can be seen as an implementation of a cooperative bargaining game~\cite{thomsonChapter35Cooperative1994}. In this game, different parties have to decide how to maximize the surplus they jointly get. In the multi-target adversarial training, we view each party as an adversary, and they negotiate to reach an agreed gradient direction that maximizes the overall robustness. 

Inspired by the bargaining game modelling for multi-target adversarial training, we first analyze the convergence property of existing methods, \ie, MAX~\cite{tramerAdversarialTrainingRobustness2019a}, MSD~\cite{mainiAdversarialRobustnessUnion2020}, and AVG~\cite{tramerAdversarialTrainingRobustness2019a}, and identify a phenomenon namely player domination. 
Specifically, it refers to the case where one player dominates the bargaining game at any time $t$, and the gradient at any time $t$ is the same as this player's gradient. 
Furthermore, we notice that under the SVM and linear model setups, player domination always occurs when using MAX and MSD, which leads to non-convergence.
Based on such theoretical results, we propose a novel mechanism that adaptively adjusts the budgets of adversaries to avoid the player domination. 
We show that with our proposed mechanism, the overall robust accuracy of MAX, AVG and MSD improves on three representative datasets. We also illustrate the performance improvement on CIFAR-$10$ in \Cref{fig:intro}. %

In this paper, we present the first theoretical analysis of the convergence of multi-target robustness on three algorithms under two models. 
Building on our theoretical results, we introduce a new method called \textbf{AdaptiveBudget}, designed to prevent the player domination phenomenon that can cause MSD and MAX to fail to converge. 
Our extensive experimental results demonstrate the superiority of our approach over previous methods.

\section{Related work}

\textbf{Adversarial Training.} Goodfellow \etal \cite{goodfellowExplainingHarnessingAdversarial2015} show that even a small perturbation in the direction of the gradient can fool deep learning models for image classification tasks. This is later extended to a multi-step attack~\cite{DBLP:conf/iclr/KurakinGB17a} called the Basic Iterative Method, now typically referred to as the PGD attack, which significantly improves the success rate of creating adversarial examples.
Since then, various variations of the PGD attack~\cite{Brendel2019AccurateRA,Li2019NATTACKLT,croceReliableEvaluationAdversarial2020} have been proposed to overcome heuristic defenses and create stronger adversaries.
To defend against these attacks, numerous defense methods~\cite{papernotDistillationDefenseAdversarial2016,kannanAdversarialLogitPairing2018,madryDeepLearningModels2018,zhangTheoreticallyPrincipledTradeoff2019,Zhang2019YouOP,Zhang2020AttacksWD,Wu2020AdversarialWP,Wang2020OnceforAllAT,Pang2020BoostingAT,Shi2020InformativeDF,Zhang2022BuildingRE,wu2022adversarial,wu2022retrievalguard} have been developed. Among these methods, the most successful defense method is adversarial training~\cite{madryDeepLearningModels2018}, which formulates the defense problem as a minimax optimization problem and has become one of the few adversarial defenses that is still robust against stronger attacks~\cite{carliniEvaluatingRobustnessNeural2017,athalyeObfuscatedGradientsGive2018,mosbach2018logit}. As a result, empirical robustness~\cite{Pang2020Mixup,pang2021bag,pmlr-v162-gao22i,zhang2022adversarial,wang2021probabilistic} has been significantly advanced over the past few decades.

\textbf{Multi-target Adversarial Training.} 
Robustness against multiple types of attacks simultaneously is closely related to our work. Schott et al. \cite{schott2018towards} use multiple variational autoencoders to construct an architecture called ``analysis by synthesis" for the MNIST dataset. Their experimental results show that even for MNIST, it is difficult to train a model that is robust to three different adversaries. Following that, Tramer and Boneh \cite{tramerAdversarialTrainingRobustness2019a} investigate the theoretical and empirical trade-offs of adversarial robustness when defending against aggregations of multiple adversaries. Their results show that a model that is robust to the $\ell_\infty$ adversary might not be able to defend against other attacks, such as $\ell_1$ and $\ell_2$ attacks, on MNIST. To alleviate this problem, they design an augmentation-based method to achieve $\ell_2$ robustness. Later, Croce and Hein \cite{croceProvableRobustnessAll2020} propose a provable adversarial defense against all $\ell_p$ norms for $p \geq 1$ using regularization methods. From a greedy search perspective, Maini et al. \cite{mainiAdversarialRobustnessUnion2020} suggest that taking the worst-case over all steepest descent directions helps achieve better performance than MAX and AVG empirically. Recently, while not studied as a defense method, Kang et al. \cite{kangTransferAdversarialRobustness2019} investigate the transferability of adversarial robustness between models trained against different perturbation models.

\section{Preliminaries}

\subsection{Problem formulation}

The goal of multi-target adversarial training is to learn a function $f_{\mathbf{w}}: \mathcal{X} \rightarrow \{-1, +1\}$ that is robust to adversarial examples generated by multiple adversaries\footnote{In our paper, we analyze the case where three adversaries are involved, \ie, $\ell_1$, $\ell_2$ and $\ell_\infty$.}, where $f_{\mathbf{w}}$ is parameterized by $\mathbf{w}$. The multi-target robust loss of $f_{\mathbf{w}}$ is defined as $\mathbb{E}_{(\mathbf{x}, y)} [\max_{\boldsymbol{\delta} \in \mathcal{B}} \ell(f_\mathbf{w}( \mathbf{x} + \boldsymbol{\delta}), y)]$,
where $\mathcal{B}=\mathcal{B}_{1}(\epsilon_{1}) \bigcup \mathcal{B}_{2}(\epsilon_{2}) \bigcup \mathcal{B}_{\infty}(\epsilon_{\infty})$, $\mathcal{B}_{p}(\epsilon) = \{\boldsymbol{\delta} : \|\boldsymbol{\delta}\|_p\leq \epsilon\}$, and $\boldsymbol{\delta}$ is the perturbation. 
In deep learning scenarios, adversarial training (AT)~\cite{madryDeepLearningModels2018} is frequently used to train a robust classifier. Previous multi-target adversarial training work, \eg, MSD~\cite{mainiAdversarialRobustnessUnion2020}, MAX~\cite{tramerAdversarialTrainingRobustness2019a}, and AVG~\cite{tramerAdversarialTrainingRobustness2019a}, employ the following minimax objective to update the model
\begin{equation}\label{problem: deep problem}
    \min_{\mathbf{w}} \mathbb{E}_{(\mathbf{x}, y)} \max_{\boldsymbol{\delta} \in \mathcal{B}} \ell(f_\mathbf{w}(\mathbf{x} + \boldsymbol{\delta}), y)\,.
\end{equation}

\begin{algorithm}[t!]
\caption{MAX, AVG and MSD algorithms}\label{alg:max avg and msd}
\begin{algorithmic}[1]
\State \textbf{MAX}$(\text{input data } \mathbf{x}, \text{steps } k , \text{stepsize } \eta, $ perturbation budgets $ (\epsilon_\infty, \epsilon_1, \epsilon_2), \text{loss function } \ell, \text{model } f_\mathbf{w})$:
\Indent
    \State $\boldsymbol{\delta}_p \gets \operatorname{PGD}(\mathbf{x}, k , \eta, \epsilon_p, \ell, f_\mathbf{w}), \forall p \in \{1,2,\infty\}$;
    \State \textbf{Return} $\operatorname{argmax}_{\boldsymbol{\delta} \in \{\boldsymbol{\delta}_1,\boldsymbol{\delta}_2,\boldsymbol{\delta}_\infty\}} \ell(f_\mathbf{w}(\mathbf{x} + \boldsymbol{\delta}_p), y)$.
\EndIndent
\\
\State \textbf{AVG}$(\text{input data }\mathbf{x}, \text{steps } k , \text{stepsize } \eta, $ perturbation budgets $ (\epsilon_\infty, \epsilon_1, \epsilon_2), \text{loss function } \ell, \text{model } f_\mathbf{w})$:
\Indent
\State \textbf{Return} $\{\operatorname{PGD}(\mathbf{x}, k , \eta, \epsilon_p, \ell, f_\mathbf{w})\}_{p \in \{1,2,\infty\}}$. %
\EndIndent
\\
\State \textbf{MSD}$(\text{input data }\mathbf{x}, \text{steps } k , \text{stepsize } \eta, $ perturbation budgets $ (\epsilon_\infty, \epsilon_1, \epsilon_2), \text{loss function } \ell, \text{model } f_\mathbf{w})$:
\Indent
\State $ \boldsymbol{\delta}^0 = \mathbf{0}$;
\For {$i \in [k]$}
    \State $\boldsymbol{\delta}_p^{i} \gets \operatorname{PGD}_{\text{Step}}(\mathbf{x}, \boldsymbol{\delta}^{i}, \eta, \epsilon_p, \ell, f_\mathbf{w}), \forall p \in \{1,2,\infty\}$;
    \State $\boldsymbol{\delta}^{i + 1} \gets \operatorname{argmax}_{\boldsymbol{\delta}_p \in \{\boldsymbol{\delta}_1^{i},\boldsymbol{\delta}_2^{i},\boldsymbol{\delta}_\infty^{i} \}} \ell(f_\mathbf{w}(\mathbf{x} + \boldsymbol{\delta}_p^{i}), y)$;
\EndFor
\State \textbf{Return} $\boldsymbol{\delta}^{k}$.
\EndIndent
\end{algorithmic}
\end{algorithm}

\comment{
\begin{algorithm}[ht!]
\caption{PGD and PGD Step}\label{alg:PGD and PGD Step}
\begin{algorithmic}[1]
\State \textbf{PGD}$(\text{input data }\mathbf{x}, \text{steps } k , \text{stepsize } \eta, \text{distance metric } \ell_p,\text{perturbation budget } \epsilon_p, \text{loss function } \ell)$:
\Indent
\State Initialize $\boldsymbol{\delta}^0$
\For {Step $i$ in $[k]$}
\State $\boldsymbol{\delta}^{i} = $\textbf{PGD\textsubscript{Step}}$(\mathbf{x}, \boldsymbol{\delta}^{i-1}, \eta, \ell_p, \epsilon_p, \ell)$
\EndFor
\State return $\boldsymbol{\delta}^{k}$
\EndIndent
\\
\State \textbf{PGD\textsubscript{Step}}$(\text{input data }\mathbf{x}, \text{current perturbation } \boldsymbol{\delta}, \text{stepsize } \eta, \text{distance metric } \ell_p, \text{perturbation budget } \epsilon_p$, loss function $ \ell)$:
\Indent
\State Return $\operatorname{Proj}_{\operatorname{Ball}_p(0, \epsilon_p)}(\boldsymbol{\delta} + \eta \operatorname{sign} (\nabla \ell (\mathbf{x} + \boldsymbol{\delta}, y)))$
\EndIndent
\end{algorithmic}
\end{algorithm}
}

This minimax problem is  usually decomposed into a two-stage problem with a maximization problem of finding the optimal $\boldsymbol{\delta}$ and a minimization problem of finding the optimal $\mathbf{w}$ given optimal $\boldsymbol{\delta}$, and then iteratively optimizing $\boldsymbol{\delta}$ and $\mathbf{w}$ for several rounds. 
\comment{
as shown below
\begin{itemize}
    \item[1.] Algorithm gets the optimal perturbation $\boldsymbol{\delta}^t$ for the $t$-th round as
    \begin{equation}\label{problem: get delta}
        \boldsymbol{\delta}^{t}(\mathbf{x}) = \operatorname{argmax}_{\boldsymbol{\delta} \in \mathcal{B}} \ell(f_{\mathbf{w}^{t-1}}(\mathbf{x}+\boldsymbol{\delta}), y)\,.
    \end{equation}
    \item[2.] Algorithm gets the optimal parameter $w^t$ for the $t$-th round as
    \begin{equation*}
        \mathbf{w}^{t} = \operatorname{argmin}_{\mathbf{w}} \mathbb{E}_{(\mathbf{x}, y)} \ell(f_\mathbf{w}(\mathbf{x}+\boldsymbol{\delta}^{t}(\mathbf{x})), y)\,.
    \end{equation*}
\end{itemize}
}
Under the non-convex scenario, to find the approximate optimal perturbation $\boldsymbol{\delta}$ and the approximate optimal parameter $\w$, gradient descent algorithm~\cite{kingmaAdamMethodStochastic2015,duchiAdaptiveSubgradientMethods2011} and projected gradient descent (PGD) attack are used. 
Specifically, PGD runs several predefined steps as $\operatorname{PGD}_{\text{step}} (\mathbf{x}, \boldsymbol{\delta}^{i}, \eta, \ell, \epsilon_p, f_\mathbf{w}) = \operatorname{Proj}_{
\mathcal{B}_{p}(\epsilon_p)
}(\boldsymbol{\delta} + \eta \operatorname{sign} (\ell^{'} (f_\mathbf{w} (\mathbf{x} + \boldsymbol{\delta}^{i}), y)))$ to approximately find a worst-case adversarial example, where $\ell^{'} (f_\mathbf{w} (\mathbf{x} + \boldsymbol{\delta}^{i}), y)$ is the gradient of $\ell (f_\mathbf{w} (\mathbf{x} + \boldsymbol{\delta}^{i}), y)$ and $\operatorname{sign} (\cdot)$ is the sign function.

Tramer and Boneh \cite{tramerAdversarialTrainingRobustness2019a} first proposed to solve the inner maximization problem of the problem~(\Cref{problem: deep problem}), 
by the MAX (the worst-case perturbation, \cref{alg:max avg and msd}) and AVG (the augmentation of all perturbations, \cref{alg:max avg and msd}). Now, the overall minimax objective becomes as below\footnote{Here we omit most of the parameter of \textbf{MSD}, \textbf{AVG}, and \textbf{MAX} for the convenience of reading without compromising the important information.}
\begin{align*}
    &\text{MAX: } \min_{\mathbf{w}} \mathbb{E}_{(\mathbf{x}, y)} \ell(f_\mathbf{w}(\mathbf{x}+ \operatorname{\textbf{MAX}}(\mathbf{x})), y),\quad\\
    &\text{AVG: } \min_{\mathbf{w}} \mathbb{E}_{(\mathbf{x}, y)} 
    \sum_{\boldsymbol{\delta} \in \textbf{AVG}(\mathbf{x})} \ell(f_\mathbf{w}(\mathbf{x} + \boldsymbol{\delta}), y)\,.
\end{align*}

Later, Maini \etal \cite{mainiAdversarialRobustnessUnion2020} designed a ``greedy'' algorithm named MSD, which solves the inner maximization problem by simultaneously maximizing the worst-case loss overall perturbation models at each projected steepest descent step as shown in \cref{alg:max avg and msd}. And then the minimax objective becomes as 
\begin{align*}
    \text{MSD: }\min_{\mathbf{w}} \mathbb{E}_{(\mathbf{x}, y)} \ell(f_\mathbf{w}(\mathbf{x} + \operatorname{\textbf{MSD}}(\mathbf{x})), y)\,.
\end{align*}

\subsection{Cooperative bargaining game}

Cooperative bargaining game~\cite{thomsonChapter35Cooperative1994} is a process in which several parties jointly decide how to share a surplus that they can jointly gain. In the cooperative bargaining game, we have $K$ players with their own utility function $u_i : \mathcal{A} \bigcup \{ \mathbf{d} \} \rightarrow \mathbb{R}$, where $\mathcal{A}$ is the set of possible agreements and $\mathbf{d}$ is the disagreement point. The feasible set of utility is defined as $\mathcal{S} = \{(u_1(\boldsymbol{\gamma}) \ldots, u_K(\boldsymbol{\gamma})) : \boldsymbol{\gamma} \in \mathcal{A}\}$. 
The goals of players are to maximize their own utility functions. $\mathcal{S}$ is assumed to be convex and compact throughout this paper while there exists a point $\boldsymbol{\gamma} \in \mathcal{A}$ satisfying  $u_i(\boldsymbol{\gamma}) > u_i(\mathbf{d}), \forall i \in [K]$ that strictly dominates the disagreement point $\mathbf{d}$, \ie, $u_i(\boldsymbol{\gamma}) > u_i(\mathbf{d}), \forall i \in [K]$, where $[K] = \{1,2,\ldots, K\}$. 

\comment{
\begin{definition}[Pareto Improvement]
A Pareto improvement is a new situation where some players will gain, and no players will lose.
\end{definition}

\begin{definition}[Pareto Dominated]
A solution is called Pareto dominated if there exists a possible Pareto improvement.
\end{definition}

\begin{definition}[Pareto Optimality]
A solution is called Pareto optimal if no change could lead to improved satisfaction for some players without some other players losing or, equivalently, if there is no scope for further Pareto improvement. In another word, the solution must not be dominated by another. 
\end{definition}
}

\comment{
For optimizing this cooperative bargaining game, back to 1953, \cite{nash1953two} had formalized a bargaining game as Nash Bargaining and presented that under certain axioms, the bargaining problem has a unique solution known as the \textit{Nash Bargaining Solution}. This solution ensures that any changes will have a negative impact on the joint surplus, which is proportionally fair. This solution satisfies the following axioms, which were later extended to multiple players in \cite{forgo1999introduction},

\textbf{Axioms 1} \textbf{ (Pareto optimality).} That solution satisfies the Pareto optimality.

\textbf{Axioms 2}\textbf{ (Independence of irrelevant alternatives).} If the set of feasible outcomes $\mathcal{S}$ is reduced, as long as the disagreement point $\mathbf{d}$ remains unchanged, and the point $\boldsymbol{\gamma}^{*}$ originally selected remains feasible, the solution of a bargaining problem does not change.

\textbf{Axioms 3} \textbf{ (Symmetry).} The solution is invariant under all exchanges of players. 

\textbf{Axioms 4} \textbf{ (Invariant to affine transformations).} For an arbitrary affine transformation $\lambda(\cdot)$, we have $\lambda(u_i(\boldsymbol{\gamma})) = u_i(\lambda(\boldsymbol{\gamma})), \forall \boldsymbol{\gamma} \in \mathcal{A}, i \in [K]$.

In the multi-target adversarial training problem, the Axioms 1 to 3 are natural as the goal of multi-target adversarial training is to obtain a model which defends multiple perturbations. As for the Axioms 4, for a $1$-lipschitz model, it holds while for most of the deep neural networks, the solution may not satisfy it. 
}

The multi-target adversarial training can be viewed as a cooperative game in which each target (perturbation) represents a player, whose utility is derived from the overall robust accuracy (defending $\ell_1$, $\ell_2$, and $\ell_\infty$ attacks simultaneously), and all the players negotiate to reach an agreed direction. 
We formalize the multi-target adversarial training problem as a bargaining game as follows. 
This bargaining game has $K$ players and for each player, they generate a data-dependent perturbation $\boldsymbol{\delta}_{k} (\mathbf{x}), \forall k \in [K]$ to complete the adversarial training. The possible agreements $\mathcal{A}$ are $\{\sum_{k \in [K]}\boldsymbol{\gamma}_k = 1, \boldsymbol{\gamma}_k \geq 0, \forall k \in [K]\}$ and 
the disagreement points will be the set $\{\boldsymbol{\gamma}_k = 1, \boldsymbol{\gamma}_j = 0, \exists k \in [K], \forall j \in [K] \backslash \{k\}\}$, where $[K] \backslash \{k\}$ is the set containing integers from $1$ to $K$ without $k$. 
We note that the agreement set $\mathcal{A}$ is compact and convex. 
$\boldsymbol{\gamma}$ is used to aggregate the gradients and decide the final update direction. Specifically, for each updates (one data point, a mini-batch or an epoch) using gradient-based algorithms, the model is updated by $\w = \w - \eta  \sum_{k \in [K] } \boldsymbol{\gamma}_k \ell^{'} (f_\mathbf{w} (\mathbf{x} + \boldsymbol{\delta}_{k}), y)$, where $\eta$ is the learning rate.

\section{Convergence analysis}

We begin this section by presenting our theoretical results based on the two commonly adopted machine learning models. Additionally, we have developed a general framework for multi-target adversarial training to avoid the player domination phenomenon that can cause the non-convergence of MAX and MSD in the next section. Our framework is inspired by our theoretical findings. 
All missing proofs are presented in \Cref{Sec: Proofs}.

\comment{
\subsection{Two learning models}
}

\subsection{Convergence analysis on SVM model}\label{Sec: Data}

Considering the binary classification setup~\cite{tsiprasRobustnessMayBe2019}, a data point $(\mathbf{x},y)$ is sampled from a distribution $\mathcal{D}$ defined by
\begin{equation*}
    \begin{aligned}
    &y \stackrel{\text { u.a.r } }{\sim} \{+1, -1\}, 
    \quad \mathbf{x}_1 = \begin{cases}
    +y,\quad \text{w.p. } p;\\
    -y,\quad \text{w.p. } 1 - p,
    \end{cases}\\
    &\mathbf{x}_2, \ldots, \mathbf{x}_{d+1} \stackrel{\text { i.i.d. } }{\sim} \mathcal{N}(\mu y,1), 
    \end{aligned}
\end{equation*}
where $\mathbf{x} = [x_1,\ldots,x_{d+1}]\in\mathbb{R}^{d+1}$, $y$ is a Rademacher random variable, and $\mathcal{N}(\mu, \sigma^2)$ is a normal distribution with mean $\mu$ and variance $\sigma^2$. 
In our setting, $p \in [0.5, 1]$. 
$x_1$ is a robust feature, while $\mathbf{x}_2, \ldots, \mathbf{x}_{d+1}$ are non-robust features that are weakly correlated with the label. Similarly, we set $\mu$ to be large enough such that a simple classifier can get a high standard accuracy ($>99$\%), \ie, $\mu \geq 1/\sqrt{d}$. 

We train a linear model %
with soft SVM loss $\ell_{\text{soft}} (y', y) = \max (0, 1-y y') $ on the data shown above
\begin{equation}\label{Eq:obj in svm1}
\begin{aligned}
    \min_{\w}&\ \mathbb{E}_{(\mathbf{x},y)\sim \mathcal{D}}\sum_{p \in \{1,2,\infty\}} \boldsymbol{\gamma}_p \ell_{\text{soft}} \left(\w^\top(\mathbf{x} + \boldsymbol{\delta}_p), y\right), \\
    &\text{s.t. } \|\w\|_2 = 1\,,
\end{aligned}
\end{equation}
where 
$\boldsymbol{\gamma} = [\boldsymbol{\gamma}_1, \boldsymbol{\gamma}_2, \boldsymbol{\gamma}_\infty]$ satisfying $\sum_{i\in\{1,2,\infty\}}\boldsymbol{\gamma}_i = 1$. 
\comment{
\begin{equation}\label{Eq: solution of perturbations}
    \begin{aligned}
        \boldsymbol{\delta}^{*}_{\infty}(\w) &= - y \epsilon_\infty \operatorname{sign} (\w),
        \quad \boldsymbol{\delta}^{*}_{1}(\w) = \frac{-y\epsilon_1 \w}{\|\w\|_{1}},\quad\\ \boldsymbol{\delta}^{*}_{2}(\w) &= \frac{-y\epsilon_2 \w}{\|\w\|_{2}}\,.
    \end{aligned}
\end{equation}
}
    
Let $\w^{t}$ and $\boldsymbol{\delta}^{t}$ be the weight vector and the perturbation at step $t$, respectively. 
The training procedures of the SVM model with AVG, MAX and MSD are illustrated as follows
\begin{enumerate}
    \item[0.] Initialize the weights with natural training, \ie, minimizing the soft-SVM loss without perturbation as
    \begin{equation}\label{Eq: Init}
        \begin{aligned}
         \w^0 &= \argmin_{\w} \mathbb{E}_{(\mathbf{x},y) \sim \mathcal{D}}\ \ell_{\text{soft}} \left(\w^\top \mathbf{x}, y\right),\\
         &\text{s.t.}\quad \| \w \|_2 = 1 \,.
        \end{aligned}
    \end{equation}
    \item Get the optimal perturbations. With the linearity property of SVM, the closed form of optimal perturbations could be calculated by $\boldsymbol{\delta}^{t}_{\infty} = - y \epsilon_\infty \operatorname{sign} (\w^{t}), \boldsymbol{\delta}^{t}_{1} = \frac{-y\epsilon_1 \w^{t}}{\|\w^{t}\|_{1}}, \boldsymbol{\delta}^{t}_{2} = \frac{-y\epsilon_2 \w^{t}}{\|\w^{t}\|_{2}}$ at time $t$.
    \item Update the weights $\w^t$ with MAX, MSD, or AVG by
    \begin{equation*}
        \begin{aligned}
            \w^{t} = &\operatorname{argmin}_{\w} \mathbb{E}_{(\mathbf{x},y)} \hspace{-0.2cm}\sum_{p \in \{1,2,\infty\}}\hspace{-0.2cm} \boldsymbol{\gamma}_p^{t}\ell_{\text{soft}} \left(\w^\top(\mathbf{x} + \boldsymbol{\delta}_p^{t}), y\right),\\
            &\text{s.t.}\quad \|\w^t\|_2 = 1\,,   
        \end{aligned}
    \end{equation*}
    
    where $\boldsymbol{\gamma}^{t} = [1/3,1/3,1/3]$ if the algorithm is AVG; $\boldsymbol{\gamma}^{t} \in \{[1,0,0], [0,1,0], [0,0,1]\}$ if the algorithm is MAX or MSD.
    \item[3.] Loop Steps 1 and 2 for predefined number of epochs or until convergence. 
\end{enumerate}

\begin{figure}[t!]
   \begin{center}
    \begin{subfigure}[b]{0.13\textwidth}
         \centering
         \includegraphics[width=1\linewidth]{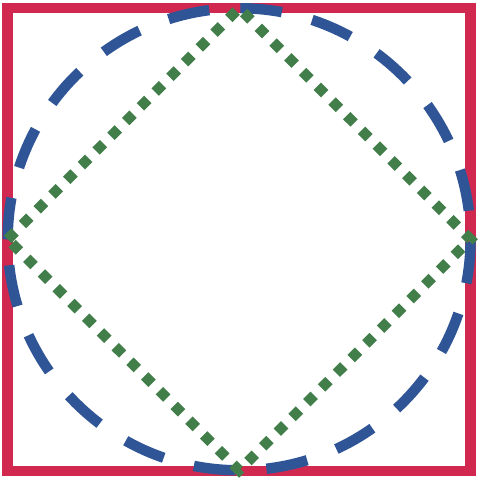}
         \caption{$d=2$}
         \label{fig:d=2}
     \end{subfigure}
     \hfill
     \begin{subfigure}[b]{0.13\textwidth}
         \centering
         \includegraphics[width=1\linewidth]{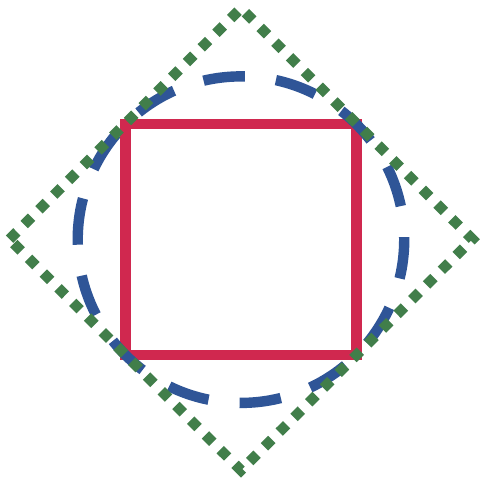}
         \caption{$d=4$}
         \label{fig:d=4}
     \end{subfigure}
     \hfill
     \begin{subfigure}[b]{0.13\textwidth}
         \centering
         \includegraphics[width=1\linewidth]{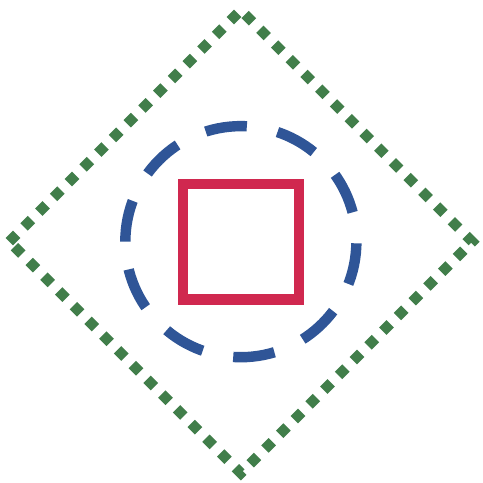}
         \caption{$d=8$}
         \label{fig:d=8}
     \end{subfigure}

      \caption{Illustration of feasible domains of $\ell_\infty$- (red region), $\ell_1$- (green dotted region), and $\ell_2$- (blue dashed region) players in $\mathbb{R}^2$, when the budgets satisfy the minimum requirements of \Cref{thm:not converge}, \ie, $\epsilon_\infty = \frac{2}{d} \epsilon_1 = \sqrt{\frac{2}{d}} \epsilon_2$. We notice that when $d=2$, the feasible regions of $1$- and $2$-players are contained in the region of $\ell_\infty$-player, while with the increase of dimension of data, the inverse case occurs and the feasible region of $\ell_\infty$-player is strictly dominated by that of $\ell_1$-player. Best view in color.
      }
      \label{fig: showing epsilons relationship}
      \vspace{-2em}
   \end{center}
\end{figure}

We first present the following negative result,
\begin{theorem}\label{thm:not converge}
Let $\mu\geq 4/\sqrt{d}$, $\epsilon_{\infty} \geq 2\mu$, $p\leq 0.977$. If one uses MAX and MSD to train the soft SVM model given $\epsilon_{\infty}
\geq \frac{2}{d}\epsilon_1$ and $\epsilon_{\infty}\geq \sqrt{\frac{2}{d}} \epsilon_2$, the loss incurred by the $\ell_\infty$-player ($\ell_\infty$-adversary) is larger than that by the $\ell_1$-player ($\ell_1$-adversary) and the $\ell_2$-player ($\ell_2$-adversary) at any time $t$ for any data sampled from the distribution $\mathcal{D}$, \ie, $\ell_{\text{soft}} \left(\w^\top(\mathbf{x} + \boldsymbol{\delta}_\infty^{t}), y\right) \geq 
\max_{p \in \{1,2\}}   \ell_{\text{soft}} \left(\w^\top(\mathbf{x} +  \boldsymbol{\delta}_p^{t}), y\right), \forall t$, $\forall (\mathbf{x}, y) \sim \mathcal{D}$.
Furthermore, $\boldsymbol{\gamma}_1=\boldsymbol{\gamma}_2=0$ and $\boldsymbol{\gamma}_\infty=1$ with MAX and MSD, which means the training dynamics of SVM model with MAX and MSD are controlled by the $\ell_\infty$-player.
\end{theorem}
\begin{remark}
    This theorem shows that even when the feasible domain of  $\ell_\infty$-adversary is much smaller than that of $\ell_1$- and $\ell_2$- adversaries (when the dimension $d$ of data is bigger than $2$), the training dynamics of SVM will still be controlled by the $\ell_\infty$-player. 
    By the definition of bargaining game in multi-target adversarial training, at any time $t$, the models update with the disagreement points. 
    As shown in \Cref{fig: showing epsilons relationship}, with the increase of dimension of the data, the feasible domain of $\ell_\infty$-adversary is strictly contained in the $\ell_1$-players' region. 
\end{remark}

We define the phenomenon where one player "dominates" the multi-target adversarial training procedure (the training procedure only depends on one player) as follows
\begin{definition}[Player dominates the cooperative game]\label{def: player donimiation}
    If $\exists i \in [k]$ such that $\boldsymbol{\gamma}_i^{t} = 1$ and $\boldsymbol{\gamma}_j^{t}=0,\forall j \in [K] / \{i\}, \forall t$, then we call that $i$-player dominates the bargaining game as models achieve the same disagreement point at any time $t$.
\end{definition}

Further, we observe that this phenomenon might lead to the \textbf{non-convergence} of SVM with MAX and MSD as the sign of weights of the model flips over time when $\ell_\infty$-player dominates the bargaining game, and given $\epsilon_{\infty} > \mu$.
\comment{
Further, %
}
\begin{theorem}\label{theorem: max and msd cannot converge}
    \labelText{\comment{
        Considering Problem~(\eqref{Eq:obj in svm1}), if the following condition holds, the training procedure using MAX and MSD cannot converge
        \begin{align}
            \epsilon_\infty & \geq \max(\epsilon_1, \epsilon_2) \text{ and } \epsilon_\infty > \eta\,.
        \end{align}
    }
    Consider Problem~(\Cref{Eq:obj in svm1}) trained with MAX and MSD. If $\ell_\infty$-player dominates the bargaining game (player domination) and $\epsilon_\infty > \mu$, the weights for the non-robust features flip over time, \ie, $\operatorname{sign}(\mathbf{w}_i^{t}) = -\operatorname{sign}(\mathbf{w}_i^{t-1}), \forall i \geq 2, \forall t$. Thus, the training procedure with MAX and MSD does not converge. 
    }{theorem: max and msd cannot converge_main}
\end{theorem}
\comment{
\begin{remark}
    \ym{We notice that the assumption of this theorem is actually quite loose as }
\end{remark}
In this way, with Lemma~\ref{lemma: condition on epsilon}, the multi-target adversarial training problem can be degraded to the single-target problems~\eqref{Eq:single obj 1} and \eqref{Eq:single obj 2}.
As the $\ell_\infty$-player dominates this game, the multi-target adversarial training problem reduces to the single-target problem (\eqref{Eq:single obj 1}). Further, with Lemma~\ref{lm:optsolu}, for non-robust feature $i$, if $\eta_\infty > \eta$, we have
$
    \operatorname{sign}(w^{t}_i) = -\operatorname{sign}(w^{t-1}_i).
$
Therefore, the training procedure does not converge. We remind that in Theorem~\ref{thm:not converge}, as long as conditions hold, the player domination will occurs and as a result, the training procedure will not converge based on Theorem~\ref{theorem: max and msd cannot converge}.
}

Although we only analyze the case when the $\ell_\infty$-player dominates the bargaining game, we notice that in situations where other players dominate this bargaining game (also known as multi-target adversarial training), with certain conditions such as $\epsilon_1 > 2\mu$, the training procedure may not converge empirically. Motivated by the negative results of the SVM model, we next test a conjecture that player domination may also lead to non-convergence in linear models.

\subsection{Player domination leads to non-convergence}\label{sec: max and msd not converge}

To test our conjecture, we introduce a linear model as follows. The linear model $f_\mathbf{w}$ is parameterized by $\mathbf{w}$ and optimized by gradient-based algorithms such as AdaGrad~\cite{duchiAdaptiveSubgradientMethods2011} or Adam~\cite{kingmaAdamMethodStochastic2015}. The parameter at time (epoch) $t$ is denoted by $\mathbf{w}^{t}$. The loss function of each player is denoted by $\ell_{k}$, where $k\in[K]$, which is $L$-smooth and $\mu$-strongly convex, and the corresponding gradient at time $t$ is denoted as $g_k^t$, where $k \in [K]$ for all $t$. We assume that for a sequence $\{\w^{t}\}_{t \in [1, \infty]}$ generated by any gradient-based optimization algorithm, the set of gradient vectors $\{g_k^t\}_{ k \in [K]}$ at any time $t$ and at any partial limit is linearly independent unless a locally optimal solution is achieved. All loss functions are differentiable, and all sub-level sets are bounded. The learning rate is denoted by $\eta$ such that $ \eta < \frac{2}{L} $. We also assume that the domain of weights is open and convex.

To generalize our theoretical results, we show that under this linear model, MAX and MSD still do not converge if one player dominates the game.
\begin{theorem}\label{thm:MAXMSD}
    Consider using MAX and MSD to train the linear model described above. If one player dominates the bargaining game throughout the game (see Definition~\ref{def: player donimiation}), then the loss of all players and the overall loss would increase as time $t$ grows. This means that the training procedure for the linear model described above does not converge.
\end{theorem}

\comment{
\section{Yihan's Proof}

\begin{equation}
    y\sim\{-1,1\},\quad x_1\sim\left\{\begin{aligned}
         y,\quad &\textrm{with prob } p;\\
        - y,\quad &\textrm{with prob } 1-p,
    \end{aligned}\right.\quad
    x_i\sim \cN(\mu y,1),\ i\geq2.
\end{equation}
Consider Madry's setting, for adversarial training we first calculate optimal $\w^{(t)}$ then calculate optimal $\boldsymbol{\delta}^{(t)}$, $t\geq 0$.
\begin{lemma}
Every entry of $\w^{(0)}$ is positive (follow Madry's proof)
\end{lemma}

\begin{lemma}
For $\ell_1,\ell_2,\ell_\infty$ attacks, if the perturb budgets satisfy $\epsilon_1= Kd\mu,\epsilon_2= K\sqrt{d}\mu,\epsilon_\infty= K\mu, K\geq2 $, then for $i\geq 2$, $sign(w^{(t)}_i) = -sign(w^{(t-1)}_i)$
\end{lemma}
\begin{proof}
Proof sketch: we consider $\ell_1$ first, the rest attack types are similar.

\textbf{Step 1.} $w^{(0)}_2 = ...=w^{(0)}_{d+1}=m,m>0$ and $w^{(0)}_1\leq m/\sqrt{d}$ (See Lemma D.1 of at odds).

\textbf{Step 2.} $\boldsymbol{\delta}^{(0)} = -\frac{y\epsilon_1\w^{(0)}}{|| \w^{(0)}||_1}$.

\textbf{Step 3.} The perturbed dataset is 
\begin{equation}
    y\sim\{-1,1\},\quad x_1\sim\left\{\begin{aligned}
         y-\frac{y\epsilon_1w^{(0)}_1}{|| \w^{(0)}||_1},\quad &\textrm{with prob } p;\\
        - y-\frac{y\epsilon_1w^{(0)}_1}{|| \w^{(0)}||_1},\quad &\textrm{with prob } 1-p,
    \end{aligned}\right.\quad
    x_i\sim \cN(\mu y-\frac{y\epsilon_1m}{|| \w^{(0)}||_1},1),\ i\geq2.
\end{equation}

\textbf{Step 4.} $dm\leq || \w^{(0)}||_1\leq m(d+1/\sqrt{d})$, $$\frac{\epsilon_1w^{(0)}_1}{|| \w^{(0)}||_1}\leq K\mu/\sqrt{d}, \textrm{(small enough and may be neglect with large d)}$$
$$\frac{\epsilon_1m}{|| \w^{(0)}||_1}\geq Kd\mu/(d+1/\sqrt{d})\geq 2\mu, \textrm{(just select a suitable K e.g. $K=2d/(d+1/\sqrt{d}))$}$$

\textbf{Step 5.} The perturbed dataset is approximately
\begin{equation}
    y\sim\{-1,1\},\quad x_1\sim\left\{\begin{aligned}
         y,\quad &\textrm{with prob } p;\\
        - y,\quad &\textrm{with prob } 1-p,
    \end{aligned}\right.\quad
    x_i\sim \cN(-\mu y,1),\ i\geq2.
\end{equation}
\end{proof}

\textbf{Step 6.} $w^{(1)}_2 = ...=w^{(1)}_{d+1}=-m'$ and $w^{(1)}_1\leq m'/\sqrt{d}$, where $m'>0$ (See Lemma D.1 of at odds).

\textbf{Step 7.} Repeat Step 1-6 we can always have for $i\geq 2$, $sign(w^{(t)}_i) = -sign(w^{(t-1)}_i)$

}

While we have shown that MAX and MSD do not converge under the two models that we study, we notice that AVG provably converges as the loss is decreasing \textit{w.r.t} the number of epochs. See the following theorem.
\begin{theorem}\label{thm:converge}
    Using AVG to train the linear model, the overall loss decreases as time $t$ grows.
\end{theorem}

This theorem shows that under the same setting, while the loss of each player and the overall loss will increase as time grows with MAX and MSD, the overall loss will decrease with AVG. The key factor that results in the non-convergence phenomenon with MAX and MSD is the player domination phenomenon, where players reach the same disagreement point all the time, leading to an increase in loss. Since AVG does not achieve any disagreement point, the player domination phenomenon does not occur, and convergence is possible. Therefore, the key to avoiding the non-convergence of MAX and MSD may be to avoid player domination, which inspires us to design the new algorithm introduced in the next section.

\comment{
\begin{definition}
    AVG-based algorithms use the linear combination of the gradients and the learning rate $\eta$ to update the model, \ie, $w^{t+1} = w^{t} - \eta \sum_{k \in [K]} \boldsymbol{\gamma}_k g_{k}^t, \sum_{k \in [K]} \boldsymbol{\gamma}_k = 1, \boldsymbol{\gamma}_k \geq 0, \forall k \in [K]$, where $\boldsymbol{\gamma}$ is generated by some algorithms, \eg, AVG, NashMTL and CAGrad.
\end{definition}

\begin{claim}
    NashMTL and CAGrad, AVG-based algorithms, converge and NashMTL converges to the Pareto optimal point if all the loss functions are convex.
\end{claim}
}

\begin{algorithm}[t!]
\caption{Framework of Multi-target Adversarial Training with \textbf{AdaptiveBudget}
}\label{alg:adaptive epsilon}
\begin{algorithmic}[1]
\Require Training epochs $E$, dataset $(\mathcal{X}, \mathcal{Y})$, adversarial budgets $(\epsilon_\infty, \epsilon_1, \epsilon_2)$, model $f(\cdot)$, loss function $\ell$. 
\For {$e \in [E]$}
    \For{$\mathbf{x}, y \in (\mathcal{X}, \mathcal{Y})$}
        \State $\boldsymbol{\delta}_p (\mathbf{x}) \gets \operatorname{PGD}(\mathbf{x}, k , \eta, \epsilon_p, \ell, f), g_p \gets \ell^{'}(f(\mathbf{x} + \boldsymbol{\delta}_{p}(\mathbf{x})), y), \forall p \in \{1,2,\infty\}$;
        \State Get adaptive budgets $\hat{\epsilon}_{1}, \hat{\epsilon}_{2}, \hat{\epsilon}_{\infty} \gets \textbf{AdaptiveBudget}$ $([g_1, g_2, g_\infty], [\epsilon_1, \epsilon_2, \epsilon_\infty])$;
        \State 
        Adversarial training
        using 
        MAX, MSD or AVG with budgets $(\hat{\epsilon}_{1}, \hat{\epsilon}_{2}, \hat{\epsilon}_{\infty})$;

    \EndFor
\EndFor
\State \textbf{Return} the classifier $f$.
\\
\State \textbf{AdaptiveBudget}$(
[g_1, g_2, g_\infty],$  
$[\epsilon_1, \epsilon_2, \epsilon_\infty])$:
\Indent
    \State $p_\text{max} \gets \argmax_{p \in \{\infty, 1, 2\}}\|g_p\|$; \quad
    \State $p_\text{min} \gets \argmin_{p \in \{\infty, 1, 2\}}\|g_p\|$;
    \State $p_\text{mid} \gets \{1,2,\infty\} / \{p_\text{max}, p_\text{min}\}$;
    \State $\epsilon_{p_\text{max}} \gets \epsilon_{p_\text{max}}\cdot\frac{\|g_{p_\text{max}}\|}{ \|g_{p_\text{mid}}\|}$,\quad %
    $\epsilon_{p_\text{min}} \gets \epsilon_{p_\text{min}} \cdot \frac{\|g_{p_\text{min}}\|}{ \|g_{p_\text{mid}}\|}$;
    \State \textbf{Return}  $\epsilon_1, \epsilon_2, \epsilon_\infty$.
\EndIndent
\end{algorithmic}
\end{algorithm}
\vspace{-0.3cm}

\section{Avoiding player domination via AdaptiveBudget}

In this section, we present the proposed algorithm \textbf{AdaptiveBudget} summarized in \Cref{alg:adaptive epsilon}.

Our theoretical results (\Cref{theorem: max and msd cannot converge} and \Cref{thm:MAXMSD}) show that MAX and MSD cannot converge when player domination occurs (Definition~\ref{def: player donimiation}).
Indeed, to achieve convergence of the model, researchers can directly use AVG~\cite{tramerAdversarialTrainingRobustness2019a} instead of MAX~\cite{tramerAdversarialTrainingRobustness2019a} and MSD~\cite{mainiAdversarialRobustnessUnion2020}.
However, previous works~\cite{tramerAdversarialTrainingRobustness2019a,mainiAdversarialRobustnessUnion2020} have shown that under the non-convex scenario, where a deep neural network with non-linear activation is trained on MNIST~\cite{lecunGradientbasedLearningApplied1998} and CIFAR-10~\cite{krizhevskyLearningMultipleLayers2009}, MSD and MAX outperform AVG\footnote{This does not conflict with our theoretical analysis as the training dynamics of non-convex and convex scenarios (\eg, SVM and linear models) are different.
Additionally, since MAX~\cite{tramerAdversarialTrainingRobustness2019a} and MSD~\cite{mainiAdversarialRobustnessUnion2020} are greedy algorithms that take steepest gradients at each time $t$, such greedy updates benefit under non-convex scenarios.}.
We have also come to a similar conclusion as shown in \Cref{tab:mnist-pgd} and \Cref{tab:cifar-pgd}.
Therefore, inspired by the previous theoretical analysis, to avoid player domination, we increase the budget of the player with the largest gradient and force the model to better handle this adversary. Intuitively, if the model can handle one adversary (player) well, the gradient of that adversary (player) will be small. So, to advance multi-target robustness, we present a novel general-purpose algorithm for multi-target adversarial robustness called \textbf{AdaptiveBudget}, which adaptively changes the budget of different adversaries to avoid the player domination phenomenon (achieving the same disagreement point).

The core idea of this algorithm is to avoid player domination by adaptively assigning proper attack budgets to different adversaries (players). Such an assignment is intended to ensure that no single player's loss is significantly larger than others, and thus alleviate player domination. In each epoch, the player who controls the updates will be different. Concretely, for each batch of data, we first obtain adversarial perturbations $\boldsymbol{\delta}_{\infty}$, $\boldsymbol{\delta}_1$, and $\boldsymbol{\delta}_2$ for the $\ell_\infty$-, $\ell_1$-, and $\ell_2$- adversaries (Step 4). Then, based on the norms ($\ell_1$ or $\ell_2$ norms) of the gradients by forwarding their adversarial examples through our model, the algorithm adaptively adjusts the budgets $\epsilon$ for different adversaries to avoid the player domination phenomenon (Step 5). Specifically, our proposed method does not change the budget of the adversary whose norm of gradient is the middle one, increases the budget of the adversary whose norm of gradient is the maximum, and decreases the budget of the adversary whose norm of gradient is the minimum. The intuition behind our method is to focus on the hardest task in the current round so that this task might be easier to model in the next round and might not be able to dominate the updates. After obtaining the adjusted adversarial budgets, the model utilizes MSD, MAX, or AVG to approximately solve the inner maximization problem and then updates its parameter with a gradient descent algorithm.

The proposed framework is general and can be applied to all existing multi-target adversarial training algorithms. The \textbf{AdaptiveBudget} module is employed to break the curse of player domination, which might occur when applying MAX and MSD to train a robust model. 
In the next section, we provide extensive experimental evidence to support the consistent effectiveness of the {AdaptiveBudget} method.

\comment{
\begin{algorithm}[t!]
\caption{Adaptive budget}\label{alg:adaptive epsilon}
\begin{algorithmic}[1]
\Require Input data $\mathbf{x}$, .
\State Get the adversarial examples by PGD, \ie, $\{\boldsymbol{\delta}_p (x) = \operatorname{PGD}(\mathbf{x}, k ,\eta, \ell_p, \epsilon_p, \ell)\}_{p \in \{1,2,\infty\}}$.
\State Get the gradients for each adversary.
\State Get the index of maximum, minimum and middle item of the gradients, \ie, $\text{maxindex} = \argmax_{p \in \{\infty, 1, 2\}}(\ell^{'}(\mathbf{x} + \boldsymbol{\delta}_p))$, $\text{minindex} = \argmin_{p \in \{\infty, 1, 2\}}(\ell^{'}(\mathbf{x} + \boldsymbol{\delta}_p))$, and $\text{midindex} = \{1,2,\infty\} / \{\text{maxindex}, \text{minindex} \}$.
\State Adjust the budget according to the norm of the gradients, \ie, $\epsilon_{\text{maxindex}} = \epsilon_{\text{maxindex}} * \|g_{\text{maxindex}}\| / \|g_{\text{midindex}}\|$ and $\epsilon_{\text{minindex}} = \epsilon_{\text{minindex}} * \|g_{\text{minindex}}\| / \|g_{\text{midindex}}\|$.
\State Utilize the adjusted budgets $(\epsilon_\infty, \epsilon_1, \epsilon_2)$ to update the model with MAX, MSD or AVG.
\end{algorithmic}
\end{algorithm}
}

\section{Experiments}

\subsection{Experimental setup and implementation details}

\textbf{Datasets.} We conducted extensive experiments on 
one synthetic dataset (\cref{Sec: Data}) to complement our theoretical results, and on MNIST~\cite{lecunGradientbasedLearningApplied1998}, CIFAR-10~\cite{krizhevskyLearningMultipleLayers2009}, and CIFAR-100~\cite{krizhevskyLearningMultipleLayers2009} to show the superiority of our proposed methods over the existing methods of multi-target adversarial training. Due to the limitation of space, the experiments on synthetic data is in Appendix.

\textbf{Methods.} 
Models that defend against multiple adversaries are trained using MAX~\cite{tramerAdversarialTrainingRobustness2019a}, AVG~\cite{tramerAdversarialTrainingRobustness2019a}, and MSD~\cite{mainiAdversarialRobustnessUnion2020}. 
For each algorithm, we use the default hyperparameters introduced in their original papers. 
All methods are implemented in PyTorch~\cite{paszkePyTorchImperativeStyle2019} on a single NVIDIA A100 GPU. 
Raw images are resized to $28 \times 28$ pixels for MNIST and $32 \times 32$ pixels for CIFAR-10 and CIFAR-100 as inputs. 
We apply the {AdaptiveBudget} to MAX, MSD, and AVG with $\ell_1$ and $\ell_2$ norms to assign proper budgets adaptively to avoid player domination.

\textbf{Models.} 
Following MSD~\cite{mainiAdversarialRobustnessUnion2020} and Madry \etal~\cite{madryDeepLearningModels2018}, for MNIST, we use a four-layer convolutional network which consists of two convolutional layers of 32 and 64 $5 \times 5$ filters and $2$ units of padding, followed by a fully connected layer with $1024$ hidden units, where both convolutional layers are followed by $2 \times 2$ Max Pooling layers and ReLU activations. 
Similarly, following MSD~\cite{mainiAdversarialRobustnessUnion2020}, for CIFAR-10 and CIFAR-100, we use the pre-activation version of the ResNet18~\cite{heDeepResidualLearning2016} architecture that consists of nine residual units with two convolutional layers.

\begin{table*}[t]
\centering
\caption{Summary of robust accuracy for MNIST (higher is better). ``w. AdaptiveBudget'' refers to employing AdaptiveBudget which aims to avoid any player dominating the game. 
``*'' means that the results are reproduced from the implementation of MSD~\cite{mainiAdversarialRobustnessUnion2020} with the hyperparameters introduced in MSD~\cite{mainiAdversarialRobustnessUnion2020}. ``$\ell_1$ (ours)'' and ``$\ell_2$ (ours)'' refers to employing our proposed AdaptiveBudget method w.r.t $\ell_1$ and $\ell_2$ norms. Note that multi-target robustness focuses on the \textbf{overall robust accuracy} (``\textbf{All Robust Acc}'' in the table).
}
\vspace{-0.3cm}
\label{tab:mnist-pgd}
\resizebox{\textwidth}{!}{%
\begin{tabular}{l|c|c|c|ccc|ccc|ccc}
\toprule
Models                   & \multirow{2}{*}{$\ell_1$} & \multirow{2}{*}{$\ell_2$} & \multirow{2}{*}{$\ell_\infty$} & \multicolumn{3}{c|}{MAX}                                 & \multicolumn{3}{c|}{MSD}                                 & \multicolumn{3}{c}{AVG}                                  \\
w. \textbf{AdaptiveBudget}      &          &          &                                    &      & $\ell_1$ (ours)               & $\ell_2$ (ours)                &      & $\ell_1$ (ours)                & $\ell_2$ (ours)                &      & $\ell_1$ (ours)                & $\ell_2$ (ours)                \\\midrule\midrule
Clean Accuracy ($\%$)         & 97.2*     & 99.1*     & 99.2*                               & 98.6* & \textbf{98.9}$\uparrow$                    & \textbf{98.9}$\uparrow$                    & 98.2* & \textbf{98.3}$\uparrow$                    & \textbf{98.9}$\uparrow$                    & 99.1* & 99.1                    & 99.1                    \\\midrule\midrule
$\ell_1$ PGD Robust Acc ($\%$)      & 47.3*     & 67.8*     & 54.6*                               & 67.1* & \textbf{71.4$\uparrow$} & \textbf{69.7$\uparrow$} & 67.3* & 66.8$\downarrow$        & 65.9$\downarrow$        & 70.6* & 68.2$\downarrow$        & 68.9$\downarrow$        \\\midrule
$\ell_2$ PGD Robust Acc ($\%$)      & 24.1*     & 66.8*     & 61.8*                               & 67.2* & \textbf{69.4$\uparrow$} & \textbf{69.5$\uparrow$} & 68.0* & 67.9$\downarrow$        & 65.3$\downarrow$        & 69.4* & 68.3$\downarrow$        & 68.3$\downarrow$        \\\midrule
$\ell_\infty$ PGD Robust Acc ($\%$) & 0*        & 0.1*      & 88.9*                               & 21.2* & \textbf{67.2$\uparrow$} & \textbf{67.6$\uparrow$} & 62.4* & \textbf{69.7$\uparrow$} & \textbf{69.7$\uparrow$} & 59.5* & \textbf{67.7$\uparrow$} & \textbf{65.6$\uparrow$} \\\midrule
\textbf{All PGD Robust Acc} ($\%$)           & 0*        & 0.1*      & 52.1*                               & 21.2* & \textbf{61.3$\uparrow$} & \textbf{61.4$\uparrow$} & 59.7* & \textbf{62.1$\uparrow$} & \textbf{61.0$\uparrow$} & 55.4* & \textbf{59.2$\uparrow$} & \textbf{58.2$\uparrow$} \\\bottomrule
\end{tabular}%
}
\vspace{-0.5cm}
\end{table*}

\textbf{Attacks used for training.} For MNIST, we follow the setting of three adversaries from MSD~\cite{mainiAdversarialRobustnessUnion2020}, as shown below. The $\ell_\infty$-adversary uses a step size of $\alpha = 0.01$ within a radius of $\epsilon_\infty=0.3$ for 50 iterations. The $\ell_2$-adversary uses a step size of $\alpha = 0.1$ within a radius of $\epsilon_2=2.0$ for 100 iterations, and the $\ell_1$-adversary uses a step size of $\alpha = 0.8$ within a radius of $\epsilon_1 = 10$ for 50 iterations. By default, the attack is run with two restarts: one starting with $\boldsymbol{\delta}=\mathbf{0}$, and another by randomly initializing $\boldsymbol{\delta}$ in the perturbation ball. 
Similarly, for CIFAR-10 and CIFAR-100, we follow MSD~\cite{mainiAdversarialRobustnessUnion2020}. The $\ell_\infty$-adversary uses a step size of $\alpha = 0.003$ within a radius of $\epsilon_\infty = 0.03$ for 40 iterations. The $\ell_2$-adversary uses a step size of $\alpha = 0.05$ within a radius of $\epsilon_2 = 0.5$ for 50 iterations, and the $\ell_1$-adversary uses a step size of $\alpha = 1.0$ within a radius of $\epsilon_1 = 12$ for 50 iterations.

\textbf{Attacks used for evaluation.} 
To fully understand the performance of the defense, we employ the PGD adversary and Autoattack~\cite{croceReliableEvaluationAdversarial2020}\footnote{We only consider white-box attacks based on gradients and do not use attacks based on gradient estimation, as the gradients for the standard architectures used here are readily available.} to test the effectiveness of our method. We make 10 random restarts for all results on MNIST, CIFAR-10, and CIFAR-100. The budgets for the three adversaries, \ie, $\epsilon_1$, $\epsilon_2$, and $\epsilon_\infty$, are the same as the setting during training for both datasets. However, we increase the number of iterations to $(100, 200, 100)$ for $(\ell_\infty, \ell_2, \ell_1)$ on MNIST, and to $(100, 500, 100)$ for $(\ell_\infty, \ell_2, \ell_1)$ on CIFAR-10 and CIFAR-100.

\textbf{Hyperparameter setting and tuning.} 
We did not tune any hyperparameters as our goal is to demonstrate the player domination phenomenon and propose a solution with our \textbf{AdaptiveBudget} method. 
We adopted all hyperparameters directly from MSD~\cite{mainiAdversarialRobustnessUnion2020}. Specifically, on MNIST, we used Adam~\cite{kingmaAdamMethodStochastic2015} without weight decay and a variation of the learning rate schedule from Smith~\cite{DBLP:journals/corr/abs-1803-09820}. The schedule is piecewise linear, starting from $0$ and increasing to $10^{-3}$ over the first $6$ epochs, then decreasing to $0$ over the last $9$ epochs. On CIFAR-10 and CIFAR-100, we used SGD~\cite{robbins1951stochastic} with momentum $0.9$ and weight decay $5\times 10^{-4}$ for all models. We also used a variation of the learning rate schedule from Smith~\cite{DBLP:journals/corr/abs-1803-09820} to achieve superconvergence in $50$ epochs. The schedule is piecewise linear, starting from $0$ and increasing to $0.1$ over the first $20$ epochs, then decreasing to $0.005$ over the next $20$ epochs, and finally decreasing to $0$ over the last $10$ epochs.

\textbf{Evaluation metric.} While our main target is to improve the \textbf{overall robust accuracy} on $\ell_1$-, $\ell_2$, and $\ell_\infty$- attacks, we report the single attack accuracy as well. The overall robust accuracy is calculated as $\sum_{(\mathbf{x}, y)} (\mathbf{I} (f(\mathbf{x} + \boldsymbol{\delta}_1 (\mathbf{x}) ) == y) * \mathbf{I} (f(\mathbf{x} + \boldsymbol{\delta}_2 (\mathbf{x}) ) == y) * \mathbf{I} (f(\mathbf{x} + \boldsymbol{\delta}_\infty (\mathbf{x}) ) == y)) / n$, where $\mathbf{I}(cond) = 1$ when $cond$ is true and $\mathbf{I}(cond) = 0$ when $cond$ is false, $n$ is the total number of testing data, and $f(\cdot)$ is the trained model.

\subsection{Results on MNIST}

\comment{
\begin{table}[t]
\centering
\caption{Summary of adversarial accuracy results for CIFAR-10 (higher is better). ``w. AdaptiveBudget'' refers to employ our proposed method which enables adapting epsilon to avoid the player domination phenomenon while $\ell_1$ and $\ell_2$ means using $\ell_1$ and $\ell_2$ metrics to calculate the norm of gradients. ``AA'' refers to AutoAttack.}
\label{tab:cifar-aa}
\resizebox{\textwidth}{!}{%
\begin{tabular}{l|c|c|c|ccc|ccc|ccc}
\toprule
Models                  & $\ell_1$ & $\ell_2$ & \multicolumn{1}{c|}{$\ell_\infty$} & \multicolumn{3}{c|}{MAX}                                 & \multicolumn{3}{c|}{MSD}                                 & \multicolumn{3}{c}{AVG}                                  \\ \midrule
w. AdaptiveBudget     &          &          &                                    &      & $\ell_1$ (ours)                & $\ell_2$ (ours)                &      & $\ell_1$ (ours)                & $\ell_2$ (ours)                &      & $\ell_1$ (ours)                & $\ell_2$ (ours)                \\\midrule\midrule
Clean Accuracy          & 92.4     & 87.5     & 84.2                               & 79.6 & 76.9                    & 78.7                    & 79.2 & 77.6                    & 79.0                    & 83.8 & 81.6                    & 81.5                    \\\midrule\midrule
$\ell_1$ AA Attack      &     0     &   23.8       &     6.2                               & 41.4 & \textbf{45.7$\uparrow$} & \textbf{45.5$\uparrow$} & 45.5 & \textbf{46.4$\uparrow$} & \textbf{46.7$\uparrow$} & 49.7 & \textbf{52.7$\uparrow$} & \textbf{50.8$\uparrow$} \\\midrule
$\ell_2$ AA Attack      &      0    &     63.0     &      57.4                              & 53.7 & \textbf{60.4$\uparrow$} & \textbf{63.2$\uparrow$} & 61.9 & \textbf{62.3$\uparrow$} & \textbf{62.1$\uparrow$} & 65.4 & 64.6$\downarrow$        & \textbf{65.5$\uparrow$} \\\midrule
$\ell_\infty$ AA Attack &    0      &   26.1       &       48.0                             & 38.4 & \textbf{44.7$\uparrow$} & \textbf{44.1$\uparrow$} & 43.1 & \textbf{45.2$\uparrow$} & \textbf{44.4$\uparrow$} & 37.0 & \textbf{43.1$\uparrow$} & \textbf{42.1$\uparrow$} \\\midrule
AA Attack All           &     0     &     19.5     &      6.2                              & 37.6 & \textbf{42.9$\uparrow$} & \textbf{42.3$\uparrow$} & 41.6 & \textbf{43.4$\uparrow$} & \textbf{43.0$\uparrow$} & 36.6 & \textbf{42.5$\uparrow$} & \textbf{41.2$\uparrow$} \\\bottomrule
\end{tabular}%
}
\end{table}
}

\begin{table*}[t]
\centering
\caption{Summary of robust accuracy for CIFAR-10 (higher is better). ``w. AdaptiveBudget'' refers to employing AdaptiveBudget which aims to avoid any player dominating the game. 
``AA'' refers to AutoAttack. ``*'' means that the results are reproduced from the implementation of MSD~\cite{mainiAdversarialRobustnessUnion2020} with the hyperparameters introduced in MSD~\cite{mainiAdversarialRobustnessUnion2020}. ``$\ell_1$ (ours)'' and ``$\ell_2$ (ours)'' refers to employing our proposed AdaptiveBudget method w.r.t $\ell_1$ and $\ell_2$ norms. Note that multi-target robustness focuses on the \textbf{overall robust accuracy} (``\textbf{All Robust Acc}'' in the table).
}
\label{tab:cifar-pgd}
\vspace{-0.3cm}
\resizebox{\textwidth}{!}{%
\begin{tabular}{l|c|c|c|ccc|ccc|ccc}
\toprule
Models                   & \multirow{2}{*}{$\ell_1$} & \multirow{2}{*}{$\ell_2$} & \multirow{2}{*}{$\ell_\infty$} & \multicolumn{3}{c|}{MAX}                                 & \multicolumn{3}{c|}{MSD}                                 & \multicolumn{3}{c}{AVG}                                  \\
w. \textbf{AdaptiveBudget}      &          &          &                                    &      & $\ell_1$ (ours)                & $\ell_2$ (ours)                &      & $\ell_1$ (ours)                & $\ell_2$ (ours)               &      & $\ell_1$ (ours)                & $\ell_2$ (ours)                \\\midrule\midrule
Clean Accuracy           & 92.4*     & 87.5*     & 84.2*                               & 79.6* & 76.9                    & 78.7                    & 79.2* & 77.6                    & 79.0                    & 83.8* & 81.6                    & 81.5                    \\\midrule\midrule
$\ell_1$ PGD Robust Acc (\%)      &   90.8*       &   31.7*       &     17.3*                               & 44.0* & \textbf{50.7$\uparrow$} & \textbf{51.7$\uparrow$} & 50.8* & \textbf{51.2$\uparrow$} & \textbf{52.6$\uparrow$} & 55.7* & \textbf{57.3$\uparrow$} & \textbf{56.3$\uparrow$} \\\midrule
$\ell_2$ PGD Robust Acc (\%)      &    0.1*      &     64.0*     &     60.6*                               & 55.6* & \textbf{63.4$\uparrow$} & \textbf{65.1$\uparrow$} & 64.3* & 63.6$\downarrow$        & \textbf{65.5$\uparrow$} & 67.0* & 66.6$\downarrow$        & 67.0                    \\\midrule
$\ell_\infty$ PGD Robust Acc (\%) &      0*    &     27.8*     &    51.2*                                & 41.3* & \textbf{47.5$\uparrow$} & \textbf{47.6$\uparrow$} & 45.7* & \textbf{48.4$\uparrow$} & \textbf{47.2$\uparrow$} & 39.4* & \textbf{45.5$\uparrow$} & \textbf{44.2$\uparrow$} \\\midrule
\textbf{All PGD Robust Acc} (\%)           &    0*      &    23.8*      &    17.3*                                & 40.4* & \textbf{46.0$\uparrow$} & \textbf{46.8$\uparrow$} & 44.1* & \textbf{47.2$\uparrow$} & \textbf{46.4$\uparrow$} & 39.2* & \textbf{45.2$\uparrow$} & \textbf{43.6$\uparrow$} \\\midrule\midrule
$\ell_1$ AA Robust Acc (\%)      &     0*     &   23.8*       &     6.2*                               & 41.4* & \textbf{45.7$\uparrow$} & \textbf{45.5$\uparrow$} & 45.5* & \textbf{46.4$\uparrow$} & \textbf{46.7$\uparrow$} & 49.7* & \textbf{52.7$\uparrow$} & \textbf{50.8$\uparrow$} \\\midrule
$\ell_2$ AA Robust Acc (\%)      &      0*    &     63.0*     &      57.4*                              & 53.7* & \textbf{60.4$\uparrow$} & \textbf{63.2$\uparrow$} & 61.9* & \textbf{62.3$\uparrow$} & \textbf{62.1$\uparrow$} & 65.4* & 64.6$\downarrow$        & \textbf{65.5$\uparrow$} \\\midrule
$\ell_\infty$ AA Robust Acc (\%) &    0*      &   26.1*       &       48.0*                             & 38.4* & \textbf{44.7$\uparrow$} & \textbf{44.1$\uparrow$} & 43.1* & \textbf{45.2$\uparrow$} & \textbf{44.4$\uparrow$} & 37.0* & \textbf{43.1$\uparrow$} & \textbf{42.1$\uparrow$} \\\midrule
\textbf{All AA Robust Acc} (\%)        &     0*     &     19.5*     &      6.2*                              & 37.6* & \textbf{42.9$\uparrow$} & \textbf{42.3$\uparrow$} & 41.6* & \textbf{43.4$\uparrow$} & \textbf{43.0$\uparrow$} & 36.6* & \textbf{42.5$\uparrow$} & \textbf{41.2$\uparrow$} \\\bottomrule
\end{tabular}%
}
\vspace{-0.3cm}
\end{table*}

\begin{table*}[t]
\centering
\caption{Summary of robust accuracy for CIFAR-100 (higher is better). ``w. AdaptiveBudget'' refers to employing AdaptiveBudget which aims to avoid any player dominating the game. 
``AA'' refers to AutoAttack. ``*'' means that the results are reproduced from the implementation of MSD~\cite{mainiAdversarialRobustnessUnion2020} with the hyperparameters introduced in MSD~\cite{mainiAdversarialRobustnessUnion2020}. ``$\ell_1$ (ours)'' and ``$\ell_2$ (ours)'' refers to employing our proposed AdaptiveBudget method w.r.t $\ell_1$ and $\ell_2$ norms. 
\vspace{-0.4cm}
}
\label{tab:cifar100-pgd}
\resizebox{\textwidth}{!}{%
\begin{tabular}{l|ccc|ccc|ccc}
\toprule
Models                            & \multicolumn{3}{c|}{MAX}                   & \multicolumn{3}{c|}{MSD}                   & \multicolumn{3}{c}{AVG}                   \\
w. \textbf{AdaptiveBudget}        &       & $\ell_1$ (ours) & $\ell_2$ (ours) &       & $\ell_1$ (ours) & $\ell_2$ (ours) &       & $\ell_1$ (ours) & $\ell_2$ (ours) \\
\midrule\midrule
Clean Accuracy                    & 55.49* & 56.48           & 55.53           & 56.09* & 55.52           & 54.94           & 59.94* & 57.78           & 58.16           \\\midrule\midrule
$\ell_1$ PGD Robust Acc (\%)      & 25.45* & \textbf{29.27$\uparrow$}           & \textbf{29.78$\uparrow$}           & 35.50* & 30.31$\downarrow$           & 28.87$\downarrow$           & 30.35* & \textbf{33.16$\uparrow$}           & \textbf{32.62$\uparrow$}           \\\midrule
$\ell_2$ PGD Robust Acc (\%)      & 39.55* & \textbf{40.00$\uparrow$}           & \textbf{39.85$\uparrow$}           & 40.14* & \textbf{40.28$\uparrow$}           & 39.28$\downarrow$           & 40.26* & \textbf{41.03$\uparrow$}           & \textbf{40.27$\uparrow$}           \\\midrule
$\ell_\infty$ PGD Robust Acc (\%) & 25.03* & \textbf{25.34$\uparrow$}           & \textbf{25.87$\uparrow$}           & 24.83* & \textbf{26.19$\uparrow$}           & \textbf{25.59$\uparrow$}           & 18.92* & \textbf{21.81$\uparrow$}           & \textbf{21.57$\uparrow$}           \\\midrule
\textbf{All PGD Robust Acc} (\%)  & 21.11* & \textbf{24.14$\uparrow$}           & \textbf{24.76$\uparrow$}           & 25.10* & 25.03$\downarrow$           & 24.43$\downarrow$           & 18.61* & \textbf{21.55$\uparrow$}           & \textbf{21.16$\uparrow$}           \\
\midrule\midrule
$\ell_1$ AA Robust Acc (\%)       & 13.00* & \textbf{23.00$\uparrow$}           & \textbf{20.90$\uparrow$}           & 25.10* & 24.00$\downarrow$           & 24.20$\downarrow$           & 25.20* & \textbf{28.60$\uparrow$}           & \textbf{28.00$\uparrow$}           \\\midrule
$\ell_2$ AA Robust Acc (\%)       & 36.30* & 35.60$\downarrow$           & \textbf{36.40$\uparrow$}           & 37.60* & 35.80$\downarrow$           & 36.40$\downarrow$           & 37.00* & \textbf{37.90$\uparrow$}           & \textbf{37.10$\uparrow$}           \\\midrule
$\ell_\infty$ AA Robust Acc (\%)  & 22.00* & 21.50$\downarrow$           & \textbf{22.30$\uparrow$}           & 21.80* & \textbf{22.80$\uparrow$}           & \textbf{22.70$\uparrow$}           & 16.30* & \textbf{19.00$\uparrow$}            & \textbf{19.70$\uparrow$}          \\\midrule
\textbf{All AA Robust Acc} (\%)   & 12.20* & \textbf{20.60$\uparrow$}           & \textbf{18.60$\uparrow$}           & 21.00*  & \textbf{21.30$\uparrow$}           & \textbf{21.50$\uparrow$}           & 16.10* & \textbf{18.90$\uparrow$}           & \textbf{19.50$\uparrow$}          \\\bottomrule
\end{tabular}%
}
\vspace{-0.6cm}
\end{table*}

Here we present results on the MNIST dataset, summarized in \Cref{tab:mnist-pgd}. Although it has been considered as an ``easy'' benchmark compared to CIFAR-10 or larger datasets, such as ImageNet~\cite{deng2009imagenet}, we noticed that all the single target adversarial training methods, namely $\ell_1$, $\ell_2$, and $\ell_\infty$, fail to defend against only three attacks, while the best method is $\ell_\infty$ training, which defends against almost all three attacks and outperforms the MAX method.

From \Cref{tab:mnist-pgd}, we can see that our proposed AdaptiveBudget improves the overall robust accuracy against $\ell_1$, $\ell_2$, and $\ell_\infty$ PGD attacks, as well as the $\ell_\infty$ robust accuracy for all three methods, \ie, MAX, MSD, and AVG, using both $\ell_1$ and $\ell_2$ norms. Specifically, on MAX, the $\ell_1$ and $\ell_2$ robust accuracy is improved by $4.3$\% and $2.2$\% (with $\ell_1$ norm AdaptiveBudget), $2.6\%$ and $2.3\%$ (with $\ell_2$ norm AdaptiveBudget), respectively. Additionally, we observe that our proposed method is able to avoid the player domination phenomenon even in non-convex scenarios, as 
all the robust accuracies of MAX are improved.

The all PGD robust accuracy of vanilla MAX also shows that the player domination phenomenon hinders MAX from achieving satisfactory robust accuracy for non-convex scenarios.
Maini \etal \cite{mainiAdversarialRobustnessUnion2020} and Tramer and Boneh \cite{tramerAdversarialTrainingRobustness2019a} mention that there is a trade-off between robust accuracy against $\ell_\infty$ attacks and robust accuracy against $\ell_1$ and $\ell_2$ attacks. Similar observations can be obtained from our experimental results. For MSD and AVG, the robust accuracy defending $\ell_1$ and $\ell_2$ PGD attacks drops slightly.

\textbf{Norm choice in AdaptiveBudget.} We use $\ell_1$ and $\ell_2$ norms for AdaptiveBudget, and the corresponding results are shown in Table~\ref{tab:mnist-pgd}. There is no significant difference between the experiments with $\ell_1$ and $\ell_2$ norms when using our proposed method. The differences in overall robust accuracy are only $0.1\%$, $1.1\%$, and $1.0\%$ on MAX, MSD, and AVG, respectively. The differences in separated robust accuracy are also small, which proves the generalization ability of our proposed method empirically.

\subsection{Results on CIFAR-10 and CIFAR-100}

The results are shown in \Cref{tab:cifar-pgd,tab:cifar100-pgd}, and the curve of robust accuracy on CIFAR-10 is shown in \Cref{fig: robustness curves cifar aa} in the Appendix. Due to the limitation of space, we present the most important results in the main paper while leaving the left results in the Appendix.

\textbf{Main results.} 
The results on CIFAR-10 presented in \Cref{tab:cifar-pgd} show the generalization ability of our proposed method, which improves the overall robust accuracy of PGD and AutoAttack of three methods, \ie, MSD, MAX, and AVG. We notice that the overall robust accuracy for PGD and AutoAttack is mainly restricted by how well the model defends against the $\ell_\infty$ attack.
This might be caused by the fact that the radius of the $\ell_\infty$ attack is too small compared to the radius of the $\ell_1$ and $\ell_2$ attacks, so with the updates by gradient-based algorithms, the gradient of the $\ell_\infty$ adversary is covered by the others, causing the model to ignore the $\ell_\infty$ adversary.
Furthermore, we notice that employing AdaptiveBudget with either the $\ell_1$ or $\ell_2$ norms helps models pay attention to the tasks that are not well-learned as the $\ell_\infty$ robust accuracy is relatively improved the most. For example, the $\ell_\infty$ PGD robust accuracy of MAX with AdaptiveBudget w.r.t. the $\ell_1$ norm experiences a relative $15.01\%$ improvement, while there is only a $14.03\%$ relative improvement on the $\ell_2$ PGD robust accuracy.
In addition, the trade-off between the three attacks on CIFAR-10 is different from that on MNIST. On MNIST, the $\ell_2$ robust accuracy is related to that of the $\ell_1$ adversary, while on CIFAR-10, it seems that $\ell_2$ robust accuracy is more likely to be related to $\ell_\infty$ robust accuracy.
Similar observations can be obtained on CIFAR-100 in \Cref{tab:cifar100-pgd}.

\section{Conclusion}

In this paper, to achieve the ultimate goal of robustness, \ie, defending any terms of attacks, we first formalized this problem within the context of a bargaining game and investigated the convergence properties of MAX, MSD, and AVG under two machine learning cases. 
We discovered that MAX and MSD do not converge theoretically due to a phenomenon called player domination, while AVG does not suffer from this. 
To prevent player domination during the training of robust models, we designed a novel framework for multi-target adversary training, which includes the proposed AdaptiveBudget method. 
Specifically, AdaptiveBudget adaptively changed the budget of different attacks to avoid player domination based on the norm of gradients of each adversary. 
Finally, 
we conducted experiments on three benchmarks, \ie, MNIST, CIFAR-10, and CIFAR-100. Experimental results showed that AdaptiveBudget improved the overall robust accuracy on three benchmarks, which complemented our theoretical results and also supported our finding that player domination might interfere with the training of robust models.

\section*{Acknowledgement}
This work is supported by
NSERC Discovery Grant RGPIN-2022-03215, DGECR-2022-00357.

\bibliographystyle{ieee_fullname}
\bibliography{name}

\clearpage

\appendix
\section{Ethics Statement}

Our work strongly relates to the security of machine learning, especially defending against adversarial examples. Numerous studies~\cite{DBLP:conf/iclr/KurakinGB17a,shafahi2018are,Jia2020Fooling,brendel2018decisionbased,feng2021intelligent} have shown that adversarial examples can cause modern machine learning systems to fail. To improve the robustness of machine learning models, both empirically and theoretically, various approaches~\cite{Balunovic2020Adversarial,wang2022selfensemble,tramer2018ensemble,Zheng_2020_CVPR,Chen_2021_CVPR} have been proposed. 
While the ultimate goal of robust machine learning is to defend against all possible and reasonable adversarial examples, most of the previous methods have only been proven to defend against one type of attack~\cite{kangTransferAdversarialRobustness2019,mainiAdversarialRobustnessUnion2020}. To move towards multi-target robustness, previous studies~\cite{mainiAdversarialRobustnessUnion2020,tramerAdversarialTrainingRobustness2019a} have proposed methods that ensure robustness towards $\ell_1$, $\ell_2$, and $\ell_\infty$-adversaries simultaneously. 
However, these methods are mainly empirical and have not been thoroughly explored theoretically. To address this gap, we investigated this problem theoretically and proposed a method that can improve the performance of all the previous methods. We believe that improving the overall worst-case robustness of machine learning models may lead to the ultimate goal of robustness, which is to learn a model that can defend against all types of attacks.
\section{Additional Lemmas and Proofs}\label{Sec: Proofs}
We analyze the following single-target adversarial training problem
\begin{equation}\label{Eq:single obj 1}
    \begin{aligned}
        &\min_{\w}\quad \mathbb{E}_{(x,y)\sim \mathcal{D}} \max (0, 1-y\w^\top (\x+\boldsymbol{\delta}_p)) \\
        & \text{s.t.}\quad  \|\w\|_2 = 1\,,
    \end{aligned}
\end{equation}
where $p\in\{1,2,\infty\}$ is given before the training procedure. 

\begin{lemma}[lemma D.1~\cite{tsiprasRobustnessMayBe2019}]\label{lm:odds1}
The optimal solution $\w^*=(\w_1,...,\w_{d+1})$ of our optimization problem (\Cref{Eq: Init}) must satisfy $\w_2=\ldots=\w_{d+1}$ and $\operatorname{sign}(\w_2)=\operatorname{sign}(\mu)$.
\end{lemma}

\begin{lemma}[lemma D.2~\cite{tsiprasRobustnessMayBe2019}]\label{lm:odds2}
The optimal solution $\w^*=(\w_1,\ldots,\w_{d+1})$ of our optimization problem (\Cref{Eq: Init}) must satisfy $\w_1\leq1/\sqrt{2}$ and $\w_2=\ldots=\w_{d+1}\geq1/\sqrt{2d}$.
\end{lemma}

\begin{lemma}\label{lm:optsolu}
In the adversarial training framework, for arbitrary step $t$, if $\epsilon> \mu$ and 
\begin{align*}
\small
    p\leq &1-\max\bigg(\\
    &\frac{\bbE[\max(0,1-\cN((\epsilon-\mu)\sqrt{d},1))]}{\bbE[\max(0,1+1/\sqrt{2}(1+\epsilon)-\cN((\epsilon-\mu)\sqrt{\frac{d}{2}},0.5))]},\\
    &\frac{\bbE[\max(0,1-\cN((\epsilon+\mu)\sqrt{d},1))]}{\bbE[\max(0,1+1/\sqrt{2}(1+\epsilon)-\cN((\epsilon+\mu)\sqrt{\frac{d}{2}},0.5))]}\bigg)\,,
\end{align*}
the optimal solution $\w^{t*}=(\w^t_1,\ldots,\w^t_{d+1})$ of our optimization problem must satisfy $\w^t_1\leq1/\sqrt{2}$ and $\w^t_2=\ldots=\w^t_{d+1}$ and $|\w^t_2|\geq1/\sqrt{2d}$. Moreover, $\operatorname{sign}(\w^t_i)=-\operatorname{sign}(\w^{t+1}_i), \forall i\in[2,d+1]$.
\end{lemma}
\comment{
\begin{proof}
$t=0$, by \Cref{lm:odds2} the result holds and $\operatorname{sign}(\w^0_i) = \operatorname{sign}(\mu), \forall i\in[2,d+1]$.

$t=1$, the perturbed distribution is given by
\begin{equation*}
    \begin{aligned}
        &y\sim\{-1,1\},\quad \x_1\sim\left\{\begin{aligned}
         y(1-\epsilon),\quad &\textrm{with prob } p;\\
        - y(1+\epsilon),\quad &\textrm{with prob } 1-p,
    \end{aligned}\right.\quad\\
    &\x_i\sim \mathcal{N}((\mu-\epsilon)y,1),\ i\geq 2
    \end{aligned}
\end{equation*}
Assume for the sake of contradiction that $\w_1^1\geq1/\sqrt{2}$, by \Cref{lm:odds1} we have $0\geq \w^1_2=\ldots=\w^1_{d+1} \geq-1/\sqrt{2d}$. Then, with probability at least $1 - p$, the first feature predicts the wrong label and without enough weight, the remaining features cannot compensate for it. Concretely,
\begin{equation*}
    \begin{aligned}
       &\bbE[\max(0,1-y\w^{1*\top}(\x-\boldsymbol{\delta}_\infty))]\\
       \geq& (1-p)\bbE[\max(0,1+\w_1^1(1+\epsilon)-|\w^1_2|\sum_{i=2}^{d+1}\cN(\epsilon-\mu,1))]\\
       \geq&(1-p)\bbE[\max(0,1+1/\sqrt{2}(1+\epsilon)-\cN((\epsilon-\mu)\sqrt{\frac{d}{2}},0.5))]\\
   \end{aligned}
\end{equation*}
We will now show that a solution that assigns zero weight on the first feature ($\w^1_2=1/\sqrt{d}$ and $\w_1^1 = 0$), achieves a better margin loss,
\begin{equation*}
    \begin{aligned}
       &\bbE[\max(0,1-y\w_1^{\top}(\x-\boldsymbol{\delta}_\infty))]\\
       =&\bbE[\max(0,1-\cN((\epsilon-\mu)\sqrt{d},1))]\,.
   \end{aligned}
\end{equation*}

Because $$p\leq 1-\frac{\bbE[\max(0,1-\cN((\epsilon-\mu)\sqrt{d},1))]}{\bbE[\max(0,1+1/\sqrt{2}(1+\epsilon)-\cN((\epsilon-\mu)\sqrt{\frac{d}{2}},0.5))]},$$
we have $\bbE[\max(0,1-y\w^{1*\top}(\x-\boldsymbol{\delta}_\infty))]\geq \bbE[\max(0,1-y\w^{1\top}(\x-\boldsymbol{\delta}_\infty))]$, which yields contradiction. Besides, in this case $\operatorname{sign}(\w^1_i) = \operatorname{sign}(\mu-\epsilon)=-\operatorname{sign}(\mu)=-\operatorname{sign}(\w^0_i),\forall i\in[2,d+1]$

$t=2$, the perturbed distribution is given by
\begin{equation*}
    \begin{aligned}
       &y\sim\{-1,1\},\quad \x_1\sim\left\{\begin{aligned}
         y(1-\epsilon),\quad &\textrm{with prob } p;\\
        - y(1+\epsilon),\quad &\textrm{with prob } 1-p,
    \end{aligned}\right.\quad\\
    &\x_i\sim \mathcal{N}((\mu+\epsilon)y,1),\ i\geq 2\,.
    \end{aligned}
\end{equation*}

Assume for the sake of contradiction that $\w^2_1\geq1/\sqrt{2}$, by \Cref{lm:odds1} we have $0\geq \w_2^2=\ldots=\w^2_{d+1} \geq-1/\sqrt{2d}$. Then, with probability at least $1-p$, the first feature predicts the wrong label and without enough weight, the remaining features cannot compensate for it. Concretely,
\begin{equation*}
\begin{split}
   &\bbE[\max(0,1-y\w^{2*\top}(\x-\boldsymbol{\delta}_\infty))]\\
   \geq& (1-p)\bbE[\max(0,1+\w^2_1(1+\epsilon)-|\w^2_2|\sum_{i=2}^{d+1}\cN(\epsilon+\mu,1))]\\
   \geq&(1-p)\bbE[\max(0,1+1/\sqrt{2}(1+\epsilon)-\cN((\epsilon+\mu)\sqrt{\frac{d}{2}},0.5))]\,.
   \end{split}
\end{equation*}

We will now show that a solution that assigns zero weight on the first feature ($\w_2^2=1/\sqrt{d}$ and $\w^2_1 = 0$), achieves a better margin loss.
\begin{equation*}
\begin{split}
   &\bbE[\max(0,1-y\w_2^{\top}(\x-\boldsymbol{\delta}_\infty))]\\
   =&\bbE[\max(0,1-\cN((\epsilon+\mu)\sqrt{d},1))]\,.
   \end{split}
\end{equation*}

Because $$p\leq 1-\frac{\bbE[\max(0,1-\cN((\epsilon+\mu)\sqrt{d},1))]}{\bbE[\max(0,1+1/\sqrt{2}(1+\epsilon)-\cN((\epsilon+\mu)\sqrt{\frac{d}{2}},0.5))]},$$
we have $\bbE[\max(0,1-y\w^{2*\top}(\x-\boldsymbol{\delta}_\infty))]\geq \bbE[\max(0,1-y\w^{2\top}(\x-\boldsymbol{\delta}_\infty))]$, which yields contradiction. Besides, in this case $\operatorname{sign}(\w^2_i) = \operatorname{sign}(\mu+\epsilon)=\operatorname{sign}(\mu)=-sign(\w^1_i), \forall i\in[2,d+1]$

By induction we can easily derive that $\w^t_1\leq1/\sqrt{2}$, $\w^t_2= \ldots =\w^t_{d+1}$, $|\w^t_2|\geq1/\sqrt{2d}$ and $\operatorname{sign}(\w^t_i)=-\operatorname{sign}(\w^{t+1}_i), \forall i\in[2,d+1], \forall t\geq0$.
\end{proof}
}
\begin{lemma}
If $z\sim\cN(\mu,\sigma^2)$,
\begin{align*}
    &\bbE_z[z\bbI_{z\geq 0}] =\int_0^\infty z\frac{1}{\sqrt{2\pi\sigma^2}}\exp(-\frac{(z-\mu)^2}{2\sigma^2})dz\\
    =& \frac{\sigma}{\sqrt{2\pi}}\exp(-\frac{\mu^2}{2\sigma^2})+\frac{\mu}{2}(\erf(\frac{\mu}{\sqrt{2}\sigma})+1)\,.
\end{align*}
\end{lemma}

\begin{lemma}\label{lm:constrained}
When $\mu\geq 4/\sqrt{d}$, $\epsilon \geq 2\mu$, and $p\leq 0.977$, the optimal 
solution $\w^{t*}=(\w^t_1,\ldots,\w^t_{d+1})$ of our optimization problem must satisfy $\w^t_1\leq1/\sqrt{2}$ and $\w^t_2=\ldots=\w^t_{d+1}$ and $|\w^t_2|\geq1/\sqrt{2d}$. 
\end{lemma}
\comment{
\begin{proof}
Let $\mu= m/\sqrt{d}, m\geq4$, $\epsilon = k\mu,k\geq2$.

We have,
\begin{equation*}
\begin{split}
&\frac{\bbE[\max(0,1-\cN((\epsilon-\mu)\sqrt{d},1))]}{\bbE[\max(0,1+1/\sqrt{2}(1+\epsilon)-\cN((\epsilon-\mu)\sqrt{\frac{d}{2}},0.5))]}\\
=&
\frac{\bbE[\max(0,\cN(1+m-km,1))]}{\bbE[\max(0,\cN(1+(1+m)/\sqrt{2}+km/\sqrt{2d}-km/\sqrt{2},0.5))]}\\
\leq&
\frac{\bbE[\max(0,\cN(1+m-km,1))]}{\bbE[\max(0,\cN(1+(1+m-km)/\sqrt{2},0.5))]}\,.
\end{split}
\end{equation*}

Consider the function $h(a) = \frac{\bbE[\max(0,\cN(a,1))]}{\bbE[\max(0,\cN(1+a/\sqrt{2},0.5))]} = \frac{\frac{1}{\sqrt{2\pi}}\exp(-\frac{a^2}{2})+\frac{a}{2}(\erf(\frac{a}{\sqrt{2}})+1)}{\frac{1}{2\sqrt{\pi}}\exp(-(1+\frac{a}{\sqrt{2}})^2)+\frac{1+\frac{a}{\sqrt{2}}}{2}(\erf((1+\frac{a}{\sqrt{2}}))+1)}$.

We have,
\begin{equation*}
    \begin{split}
        h'(a) = &((\frac{1}{2}+\frac{1}{2}\erf(\frac{a}{\sqrt{2}})) (\frac{1}{2\sqrt{\pi}}\exp(-(1+\frac{a}{\sqrt{2}})^2)\\
        &+\frac{1+\frac{a}{\sqrt{2}}}{2}(\erf((1+\frac{a}{\sqrt{2}}))+1))\\
        &-(\frac{1}{2\sqrt{2}}+\frac{1}{2\sqrt{2}}\erf(1+\frac{a}{\sqrt{2}}))(\frac{1}{\sqrt{2\pi}}\exp(-\frac{a^2}{2})\\
        &+\frac{a}{2}(\erf(\frac{a}{\sqrt{2}})+1))))/(\frac{1}{2\sqrt{\pi}}\exp(-a^2)\\
        &+\frac{a}{2}(\erf(a)+1))^2\,.
    \end{split}
\end{equation*}
By numerical simulation we have $h'(a)\geq 0$, when $a\leq 0$, so $h(a)$ is increasing with $a$ when $a\leq 0$, thus
\begin{align*}
    & 1-\max(\frac{\bbE[\max(0,1-\cN((\epsilon-\mu)\sqrt{d},1))]}{\bbE[\max(0,1+1/\sqrt{2}(1+\epsilon)-\cN((\epsilon-\mu)\sqrt{\frac{d}{2}},0.5))]},\\
    &\frac{\bbE[\max(0,1-\cN((\epsilon+\mu)\sqrt{d},1))]}{\bbE[\max(0,1+1/\sqrt{2}(1+\epsilon)-\cN((\epsilon+\mu)\sqrt{\frac{d}{2}},0.5))]})\\
    \geq &1-h(-3)=0.9775>p.
\end{align*}
By Lemma~\ref{lm:optsolu}, we have the optimal solution $\w^{t*}=(\w^t_1,\ldots,\w^t_{d+1})$ of our optimization problem must satisfy $\w^t_1\leq1/\sqrt{2}$ and $\w^t_2=\ldots=\w^t_{d+1}$ and $|\w^t_2|\geq1/\sqrt{2d}$. 
\end{proof}
}

\begin{lemma}\label{lm:dominate}
When $\w_t^1\leq1/\sqrt{2}$ and $\w_t^2=\ldots=\w_t^{d+1}$ and $|\w_t^2|\geq1/\sqrt{2d}$, if $\epsilon_{\infty}
\geq \frac{2}{d}\epsilon_1$ and $\epsilon_{\infty}\geq \sqrt{\frac{2}{d}} \epsilon_2$, $\infty$-player dominates $1$-player and $2$-player. In another word, the training procedure cannot converge.
\end{lemma}
\comment{
\begin{proof}
    Let $\ell_p = 1 - yw^{\top} (\x + \boldsymbol{\delta}_p)$, we have
    \begin{align*}
        \ell_\infty - \ell_1 = & y\w_t^{\top} (\boldsymbol{\delta}_1 - \boldsymbol{\delta}_\infty)= \epsilon_\infty \|\w\|_1 - \epsilon_1 \frac{\|\w_t\|_2^2}{\|\w_t\|_1}\\
        &\geq \epsilon_1(\frac{2}{d}||\w_t||_1^2-1) \\
        &\geq \epsilon_1(\frac{2}{d}(|\w_t^1|+d|\w_t^2|)^2-1) \\
        &\geq \epsilon_1(\frac{2}{d}(\frac{1}{\sqrt{2}}+d\frac{1}{\sqrt{2d}})^2-1) >0\,.,\\
        \ell_\infty - \ell_2 = & y\w^{\top} (\boldsymbol{\delta}_2 - \boldsymbol{\delta}_\infty)= \epsilon_\infty \|\w\|_1 - \epsilon_1 \frac{\|\w\|_2^2}{\|\w\|_2}\\
         &\geq \epsilon_2(\sqrt{\frac{2}{d}}||\w_t||_1-1) \\
        &\geq \epsilon_2(\sqrt{\frac{2}{d}}(|\w_t^1|+d|\w_t^2|)-1) \\
        &\geq \epsilon_2(\sqrt{\frac{2}{d}}(\frac{1}{\sqrt{2}}+d\frac{1}{\sqrt{2d}})-1) >0\,.
    \end{align*}

    Now, we have proved that $\infty$-player dominates others and $\operatorname{sign}(\w_i^{t}) = - \operatorname{sign}(\w_i^{t-1})$. With Lemma~\ref{lm:optsolu}, we know that at any time $t$, we have $|\w_{t}^i - \w_{t-1}^i| \geq \sqrt{1/d}, \forall i \in [2, d+1]$, which means the training procedure cannot converge.
\end{proof}
}

\begin{lemma}\label{lem: max=msd}
    MAX and MSD are the same under the SVM scenario.
\end{lemma}
\comment{
\begin{proof}
    Under the deep learning cases (non-linear and non-convex), MSD follows the steepest direction ($\ell_1$, $\ell_2$ or $\ell_\infty$) in each PGD step to find the perturbation which approximately maximizes the loss function, while MAX uses PGD to find the perturbations empirically and then chooses the perturbation maximizing the loss function. MSD and MAX are different approaches in deep learning cases (non-linear and non-convex). 
    
    On the other side, under the SVM (convex and linear) case, the optimal perturbations with $\ell_1$, $\ell_2$ and $\ell_\infty$ constraints have analytical solutions. 
    In this way, both MSD and MAX can directly determine which perturbation maximizes the loss within one step, which means MSD and MAX are the same under the SVM case.
\end{proof}
}

\textbf{Standard classification is easy}. Remind that the data consists of a robust feature $\mathbf{x}_1$, which is strongly related to the label and $d$ non-robust features $\mathbf{x}_i, i \in [2, d+1]$, which are weakly related to the label $y$. But with the non-robust features, we can construct a simple linear classifier $f$ that achieves over $99\%$ natural accuracy as 
\begin{equation*}
    f(\mathbf{x}) = \operatorname{sign}([0, \frac{1}{d}, \ldots, \frac{1}{d}]^{\top} \mathbf{x})\,.
\end{equation*}
For the natural accuracy, we have
\begin{align*}
    & Pr[f(\mathbf{x}) = y] = Pr[\operatorname{sign}([0, \frac{1}{d}, \ldots, \frac{1}{d}]^{\top} \mathbf{x}) = y] \\
    =& Pr[\frac{y}{d}\sum_{i=1}^{d}\mathcal{N}(\mu y, 1) > 0] = Pr[\mathcal{N}(\mu, \frac{1}{d}) > 0] \geq 0.99\,,
\end{align*}
when $\mu \geq \frac{3}{\sqrt{d}}$.

\textbf{Robust classification is not easy}. We have the opposite observation when facing $\ell_\infty$ adversarial training. The robust accuracy is shown as 
\begin{equation*}
    \begin{aligned}
        &\min_{\|\boldsymbol{\delta}_\infty\|_\infty \leq \epsilon_\infty} Pr[f(\mathbf{x + \boldsymbol{\delta}_\infty}) = y] \leq Pr[\mathcal{N}(\mu, \frac{1}{d}) - \epsilon > 0] \\
        =& Pr[\mathcal{N}(-\mu, \frac{1}{d}) > 0] \leq 0.01\,,
    \end{aligned}
\end{equation*}
when $\epsilon_\infty = 2\mu$. In the following part of our paper, we show that it is not only difficult to get a fairly good robust accuracy, but also a converged model under the multi-target adversarial training problem.

\section{Proofs}

\subsection{Proof of \Cref{thm:not converge}}
\begin{proof}
Combining  Lemma \ref{lm:optsolu}, Lemma \ref{lm:dominate}, and Lemma \ref{lem: max=msd} yields this theorem. 
\end{proof}

\subsubsection{Proof of \Cref{theorem: max and msd cannot converge}}
\begin{proof}
As the $\infty$-player dominates this game, the multi-target adversarial training problem reduces to the single-target problem~\eqref{Eq:single obj 1}. Further, with Lemma~\ref{lm:optsolu}, for non-robust feature $i$, at any time $t$, we have
    $
        \operatorname{sign}(\w^{t}_i) = -\operatorname{sign}(\w^{t-1}_i)\,.
    $
    Thus the training procedure does not converge.
\end{proof}    
\subsubsection{Proof of \Cref{thm:MAXMSD}}
\begin{proof}
    For the $i$-th player's loss (the $i$-th player dominates the bargaining game at the time $t$), as the loss function is $\mu$-strongly convex, we have
    \begin{equation*}
    \small
        \begin{aligned}
            \ell_i(\w^{t+1}) & \geq  \ell_i(\w^{t}) - \ell_i^{'}(\w^{t})^\top (\w^{t+1} - \w^{t}) + \frac{\mu}{2}\|\w^{t+1} - \w^{t}\|_2^2 \\
            \ell_i(\w^{t+1}) & \geq \ell_i(\w^{t}) + \eta \ell_i^{'}(\w^{t})^\top \ell_i^{'}(\w^{t}) + \frac{\mu\eta^2}{2} \|\ell_i^{'}(\w^{t})\|_2^2 > \ell_i(\w^{t})\,. 
        \end{aligned}
    \end{equation*}
    
    For the $j$-th player's loss and $j \neq i$, as the loss function is $\mu$-strongly convex, we have
    \begin{equation*}
    \small
        \begin{aligned}
            \ell_j(\w^{t+1}) & \geq  \ell_j(\w^{t}) - \ell_j^{'}(\w^{t})^\top (\w^{t+1} - \w^{t}) + \frac{\mu}{2}\|\w^{t+1} - \w^{t}\|_2^2 \\
            \ell_j(\w^{t+1}) & \geq \ell_j(\w^{t}) + \frac{\mu\eta^2}{2} \|\ell_i^{'}(\w^{t})\|_2^2 > \ell_j(\w^{t})\,.%
        \end{aligned}
    \end{equation*}
    
    That means at time $t$, the loss of all player will keep increasing. And thus, if one player dominate the bargaining game throughout the whole game, the loss of all players and  will keep increasing during the whole game, which means the bargaining game might not converge.
\end{proof}
\subsubsection{Proof of \Cref{thm:converge}}
\begin{proof}
    As the loss function is $L$-smooth, $\forall i$, we have
    \begin{equation*}
        \begin{aligned}
            \ell_i(\w^{t+1}) \leq & \ell_i(\w^t) + \eta \ell_i(\w^{t})^{\top} (\w^{t+1} - \w^{t}) \\
            &+ \frac{L}{2} \|\eta \sum_{k \in [K]} g_k^t / K\|^2, \ \ \ \ (\text{as L-smooth})\\
            \ell_i^{t+1} \leq & \ell_i^t - \eta g_i^{t\top} \sum_{k \in [K]} g_k^t / K + \frac{L}{2} \|\eta \sum_{k \in [K]} g_k^t / K\|^2,\\
            = & \ell_i^t - \eta g_i^{t\top} g_i / K + \frac{L \eta^2 }{2K^2} \sum_{k\in[K]} g_k^{t\top} g_k\,.
        \end{aligned}
    \end{equation*}

    Summing the above inequality from $i=1$ to $K$, we have
    \begin{equation}
        \begin{aligned}
            &\ell^{t+1} \leq & & \ell^t - \frac{\eta}{K}\sum_{k \in [K]} g_k^{t\top} g_k  + \frac{L \eta^2 }{2K} \sum_{k\in[K]} g_k^{t\top} g_k < \ell^{t}\,. \\
            &\ \ \ (\text{as } \eta < \frac{2}{L})
        \end{aligned}
    \end{equation}
    The proof is completed.
\end{proof}

\subsection{Proof of \cref{lm:optsolu}}
$t=0$, by \Cref{lm:odds2} the result holds and $\operatorname{sign}(\w^0_i) = \operatorname{sign}(\mu), \forall i\in[2,d+1]$.

$t=1$, the perturbed distribution is given by
\begin{equation*}
    \begin{aligned}
        &y\sim\{-1,1\},\quad \x_1\sim\left\{\begin{aligned}
         y(1-\epsilon),\quad &\textrm{with prob } p;\\
        - y(1+\epsilon),\quad &\textrm{with prob } 1-p,
    \end{aligned}\right.\quad\\
    &\x_i\sim \mathcal{N}((\mu-\epsilon)y,1),\ i\geq 2
    \end{aligned}
\end{equation*}
Assume for the sake of contradiction that $\w_1^1\geq1/\sqrt{2}$, by \Cref{lm:odds1} we have $0\geq \w^1_2=\ldots=\w^1_{d+1} \geq-1/\sqrt{2d}$. Then, with probability at least $1 - p$, the first feature predicts the wrong label and without enough weight, the remaining features cannot compensate for it. Concretely,
\begin{equation*}
    \begin{aligned}
       &\bbE[\max(0,1-y\w^{1*\top}(\x-\boldsymbol{\delta}_\infty))]\\
       \geq& (1-p)\bbE[\max(0,1+\w_1^1(1+\epsilon)-|\w^1_2|\sum_{i=2}^{d+1}\cN(\epsilon-\mu,1))]\\
       \geq&(1-p)\bbE[\max(0,1+1/\sqrt{2}(1+\epsilon)-\cN((\epsilon-\mu)\sqrt{\frac{d}{2}},0.5))]\\
   \end{aligned}
\end{equation*}
We will now show that a solution that assigns zero weight on the first feature ($\w^1_2=1/\sqrt{d}$ and $\w_1^1 = 0$), achieves a better margin loss,
\begin{equation*}
    \begin{aligned}
       &\bbE[\max(0,1-y\w_1^{\top}(\x-\boldsymbol{\delta}_\infty))]\\
       =&\bbE[\max(0,1-\cN((\epsilon-\mu)\sqrt{d},1))]\,.
   \end{aligned}
\end{equation*}

Because $$p\leq 1-\frac{\bbE[\max(0,1-\cN((\epsilon-\mu)\sqrt{d},1))]}{\bbE[\max(0,1+1/\sqrt{2}(1+\epsilon)-\cN((\epsilon-\mu)\sqrt{\frac{d}{2}},0.5))]},$$
we have $\bbE[\max(0,1-y\w^{1*\top}(\x-\boldsymbol{\delta}_\infty))]\geq \bbE[\max(0,1-y\w^{1\top}(\x-\boldsymbol{\delta}_\infty))]$, which yields contradiction. Besides, in this case $\operatorname{sign}(\w^1_i) = \operatorname{sign}(\mu-\epsilon)=-\operatorname{sign}(\mu)=-\operatorname{sign}(\w^0_i),\forall i\in[2,d+1]$

$t=2$, the perturbed distribution is given by
\begin{equation*}
    \begin{aligned}
       &y\sim\{-1,1\},\quad \x_1\sim\left\{\begin{aligned}
         y(1-\epsilon),\quad &\textrm{with prob } p;\\
        - y(1+\epsilon),\quad &\textrm{with prob } 1-p,
    \end{aligned}\right.\quad\\
    &\x_i\sim \mathcal{N}((\mu+\epsilon)y,1),\ i\geq 2\,.
    \end{aligned}
\end{equation*}

Assume for the sake of contradiction that $\w^2_1\geq1/\sqrt{2}$, by \Cref{lm:odds1} we have $0\geq \w_2^2=\ldots=\w^2_{d+1} \geq-1/\sqrt{2d}$. Then, with probability at least $1-p$, the first feature predicts the wrong label and without enough weight, the remaining features cannot compensate for it. Concretely,
\begin{equation*}
\begin{split}
   &\bbE[\max(0,1-y\w^{2*\top}(\x-\boldsymbol{\delta}_\infty))]\\
   \geq& (1-p)\bbE[\max(0,1+\w^2_1(1+\epsilon)-|\w^2_2|\sum_{i=2}^{d+1}\cN(\epsilon+\mu,1))]\\
   \geq&(1-p)\bbE[\max(0,1+1/\sqrt{2}(1+\epsilon)-\cN((\epsilon+\mu)\sqrt{\frac{d}{2}},0.5))]\,.
   \end{split}
\end{equation*}

We will now show that a solution that assigns zero weight on the first feature ($\w_2^2=1/\sqrt{d}$ and $\w^2_1 = 0$), achieves a better margin loss.
\begin{equation*}
\begin{split}
   &\bbE[\max(0,1-y\w_2^{\top}(\x-\boldsymbol{\delta}_\infty))]\\
   =&\bbE[\max(0,1-\cN((\epsilon+\mu)\sqrt{d},1))]\,.
   \end{split}
\end{equation*}

Because $$p\leq 1-\frac{\bbE[\max(0,1-\cN((\epsilon+\mu)\sqrt{d},1))]}{\bbE[\max(0,1+1/\sqrt{2}(1+\epsilon)-\cN((\epsilon+\mu)\sqrt{\frac{d}{2}},0.5))]},$$
we have $\bbE[\max(0,1-y\w^{2*\top}(\x-\boldsymbol{\delta}_\infty))]\geq \bbE[\max(0,1-y\w^{2\top}(\x-\boldsymbol{\delta}_\infty))]$, which yields contradiction. Besides, in this case $\operatorname{sign}(\w^2_i) = \operatorname{sign}(\mu+\epsilon)=\operatorname{sign}(\mu)=-sign(\w^1_i), \forall i\in[2,d+1]$

By induction we can easily derive that $\w^t_1\leq1/\sqrt{2}$, $\w^t_2= \ldots =\w^t_{d+1}$, $|\w^t_2|\geq1/\sqrt{2d}$ and $\operatorname{sign}(\w^t_i)=-\operatorname{sign}(\w^{t+1}_i), \forall i\in[2,d+1], \forall t\geq0$.

\subsection{Proof of Lemma~\ref{lm:constrained}}
Let $\mu= m/\sqrt{d}, m\geq4$, $\epsilon = k\mu,k\geq2$.

We have,
\begin{equation*}
\begin{split}
&\frac{\bbE[\max(0,1-\cN((\epsilon-\mu)\sqrt{d},1))]}{\bbE[\max(0,1+1/\sqrt{2}(1+\epsilon)-\cN((\epsilon-\mu)\sqrt{\frac{d}{2}},0.5))]}\\
=&
\frac{\bbE[\max(0,\cN(1+m-km,1))]}{\bbE[\max(0,\cN(1+(1+m)/\sqrt{2}+km/\sqrt{2d}-km/\sqrt{2},0.5))]}\\
\leq&
\frac{\bbE[\max(0,\cN(1+m-km,1))]}{\bbE[\max(0,\cN(1+(1+m-km)/\sqrt{2},0.5))]}\,.
\end{split}
\end{equation*}

Consider the function $h(a) = \frac{\bbE[\max(0,\cN(a,1))]}{\bbE[\max(0,\cN(1+a/\sqrt{2},0.5))]} = \frac{\frac{1}{\sqrt{2\pi}}\exp(-\frac{a^2}{2})+\frac{a}{2}(\erf(\frac{a}{\sqrt{2}})+1)}{\frac{1}{2\sqrt{\pi}}\exp(-(1+\frac{a}{\sqrt{2}})^2)+\frac{1+\frac{a}{\sqrt{2}}}{2}(\erf((1+\frac{a}{\sqrt{2}}))+1)}$.

We have,
\begin{equation*}
    \begin{split}
        h'(a) = &((\frac{1}{2}+\frac{1}{2}\erf(\frac{a}{\sqrt{2}})) (\frac{1}{2\sqrt{\pi}}\exp(-(1+\frac{a}{\sqrt{2}})^2)\\
        &+\frac{1+\frac{a}{\sqrt{2}}}{2}(\erf((1+\frac{a}{\sqrt{2}}))+1))\\
        &-(\frac{1}{2\sqrt{2}}+\frac{1}{2\sqrt{2}}\erf(1+\frac{a}{\sqrt{2}}))(\frac{1}{\sqrt{2\pi}}\exp(-\frac{a^2}{2})\\
        &+\frac{a}{2}(\erf(\frac{a}{\sqrt{2}})+1))))/(\frac{1}{2\sqrt{\pi}}\exp(-a^2)\\
        &+\frac{a}{2}(\erf(a)+1))^2\,.
    \end{split}
\end{equation*}
By numerical simulation we have $h'(a)\geq 0$, when $a\leq 0$, so $h(a)$ is increasing with $a$ when $a\leq 0$, thus
\begin{align*}
    & 1-\max(\frac{\bbE[\max(0,1-\cN((\epsilon-\mu)\sqrt{d},1))]}{\bbE[\max(0,1+1/\sqrt{2}(1+\epsilon)-\cN((\epsilon-\mu)\sqrt{\frac{d}{2}},0.5))]},\\
    &\frac{\bbE[\max(0,1-\cN((\epsilon+\mu)\sqrt{d},1))]}{\bbE[\max(0,1+1/\sqrt{2}(1+\epsilon)-\cN((\epsilon+\mu)\sqrt{\frac{d}{2}},0.5))]})\\
    \geq &1-h(-3)=0.9775>p.
\end{align*}
By Lemma~\ref{lm:optsolu}, we have the optimal solution $\w^{t*}=(\w^t_1,\ldots,\w^t_{d+1})$ of our optimization problem must satisfy $\w^t_1\leq1/\sqrt{2}$ and $\w^t_2=\ldots=\w^t_{d+1}$ and $|\w^t_2|\geq1/\sqrt{2d}$. 

\subsection{Proof of Lemma~\ref{lm:dominate}}
    Let $\ell_p = 1 - yw^{\top} (\x + \boldsymbol{\delta}_p)$, we have
    \begin{align*}
        \ell_\infty - \ell_1 = & y\w_t^{\top} (\boldsymbol{\delta}_1 - \boldsymbol{\delta}_\infty)= \epsilon_\infty \|\w\|_1 - \epsilon_1 \frac{\|\w_t\|_2^2}{\|\w_t\|_1}\\
        &\geq \epsilon_1(\frac{2}{d}||\w_t||_1^2-1) \\
        &\geq \epsilon_1(\frac{2}{d}(|\w_t^1|+d|\w_t^2|)^2-1) \\
        &\geq \epsilon_1(\frac{2}{d}(\frac{1}{\sqrt{2}}+d\frac{1}{\sqrt{2d}})^2-1) >0\,.,\\
        \ell_\infty - \ell_2 = & y\w^{\top} (\boldsymbol{\delta}_2 - \boldsymbol{\delta}_\infty)= \epsilon_\infty \|\w\|_1 - \epsilon_1 \frac{\|\w\|_2^2}{\|\w\|_2}\\
         &\geq \epsilon_2(\sqrt{\frac{2}{d}}||\w_t||_1-1) \\
        &\geq \epsilon_2(\sqrt{\frac{2}{d}}(|\w_t^1|+d|\w_t^2|)-1) \\
        &\geq \epsilon_2(\sqrt{\frac{2}{d}}(\frac{1}{\sqrt{2}}+d\frac{1}{\sqrt{2d}})-1) >0\,.
    \end{align*}

    Now, we have proved that $\infty$-player dominates others and $\operatorname{sign}(\w_i^{t}) = - \operatorname{sign}(\w_i^{t-1})$. With Lemma~\ref{lm:optsolu}, we know that at any time $t$, we have $|\w_{t}^i - \w_{t-1}^i| \geq \sqrt{1/d}, \forall i \in [2, d+1]$, which means the training procedure cannot converge.

\subsection{Proof of Lemma~\ref{lem: max=msd}}

Under the deep learning cases (non-linear and non-convex), MSD follows the steepest direction ($\ell_1$, $\ell_2$ or $\ell_\infty$) in each PGD step to find the perturbation which approximately maximizes the loss function, while MAX uses PGD to find the perturbations empirically and then chooses the perturbation maximizing the loss function. MSD and MAX are different approaches in deep learning cases (non-linear and non-convex). 
    
    On the other side, under the SVM (convex and linear) case, the optimal perturbations with $\ell_1$, $\ell_2$ and $\ell_\infty$ constraints have analytical solutions. 
    In this way, both MSD and MAX can directly determine which perturbation maximizes the loss within one step, which means MSD and MAX are the same under the SVM case.
\section{Extra Experiments}

Due to the limitation of space, we put the experimental verification of our negative results and Figure~\ref{fig: robustness curves cifar aa} here.

\subsection{Verifying effects of player domination on the SVM case}
\begin{figure*}[h]
    \begin{center}
        \begin{subfigure}{0.45\linewidth}
            \includegraphics[width=1\linewidth]{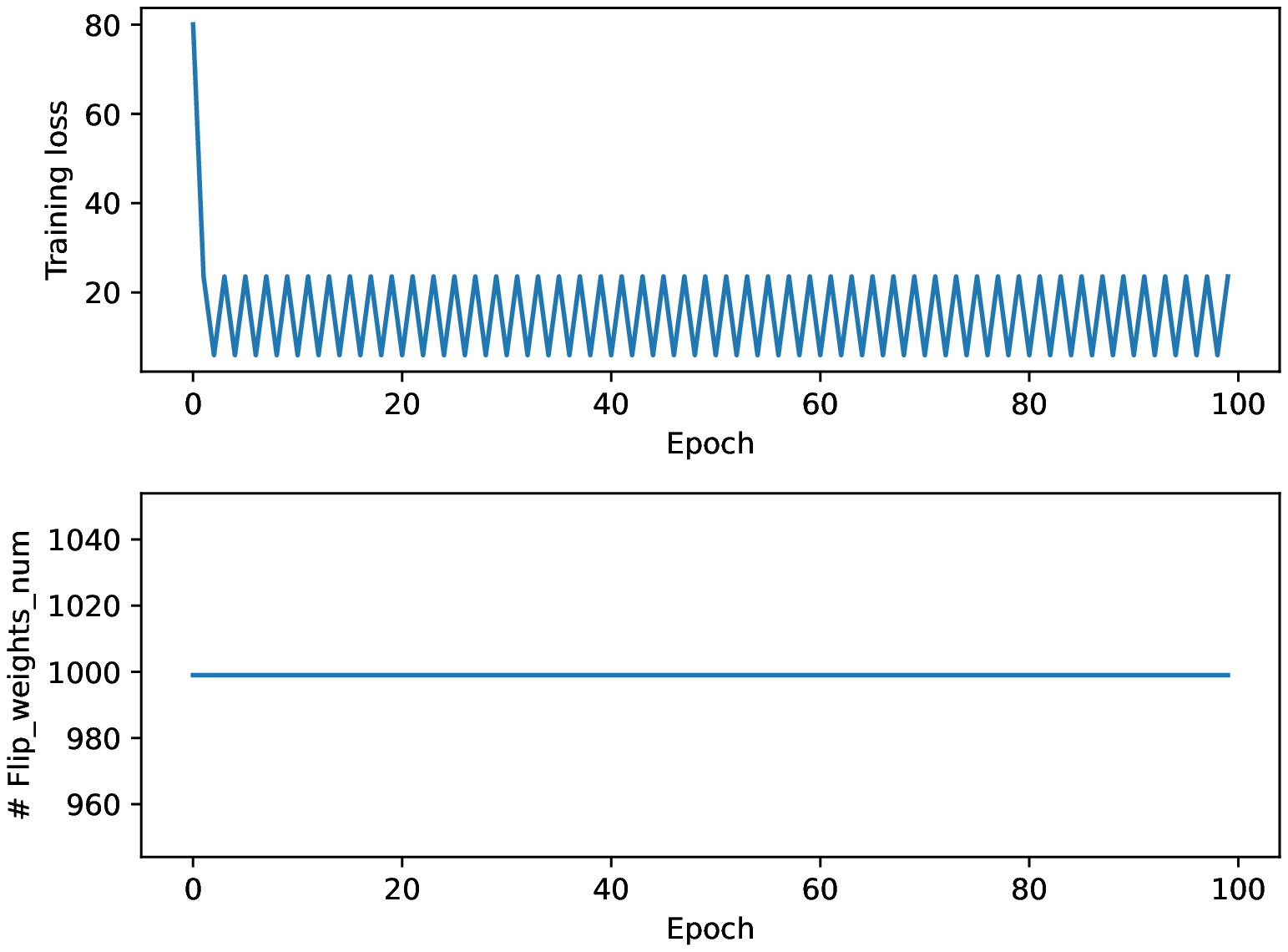}
            \caption{MAX and MSD}\label{fig: intro1_MAX MSD}
        \end{subfigure}
        \begin{subfigure}{0.45\linewidth}
            \includegraphics[width=1\linewidth]{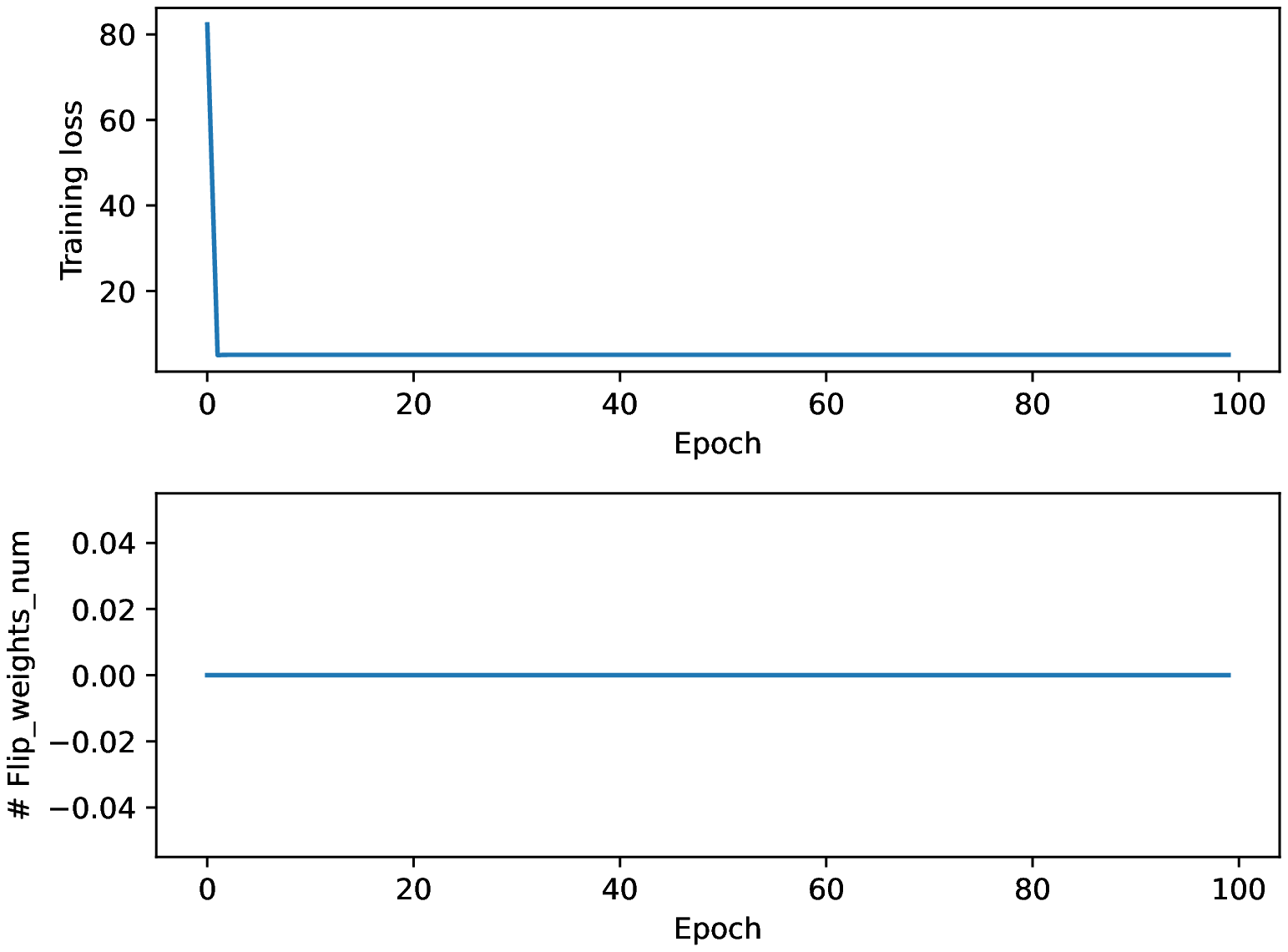}
            \caption{AVG}\label{fig: intro1_AVG}
        \end{subfigure}
        
        \caption{We illustrate the training loss and the number of weights that flip between two epochs. \Cref{fig: intro1_MAX MSD} shows the data of model trained with MAX and MSD using the SVM model (Sec~\ref{Sec: Data}) while \Cref{fig: intro1_AVG} shows the number of model trained with AVG.}
        \label{fig:intro1}
    \end{center}
\end{figure*}

To verify our theoretical results, we conduct experiments and the corresponding results are shown in Figure~\ref{fig:intro1}. We use a fully connected network (Fully Connected Layer (in=$d$, out=$1$), where $d = 1000$). For the data generation, we set $p=0.95$, $\mu = 4 / \sqrt{d}$, and $\epsilon_1=\epsilon_2=\epsilon_\infty=2\mu$, and the sample size is $100000$.

We notice that with MAX or MSD (they are equal under the SVM scenario, Lemma~\ref{lem: max=msd}), the training procedure cannot converge as the training loss is fluctuating while the number of weights whose signs are flipped compared with last epoch is almost 1000. At the same time, AVG does converge. That complements our theoretical results (Theorem~\ref{thm:not converge}).

\begin{figure*}[h]
    \begin{center}
        \begin{subfigure}{0.45\linewidth}
            \includegraphics[width=1\linewidth]{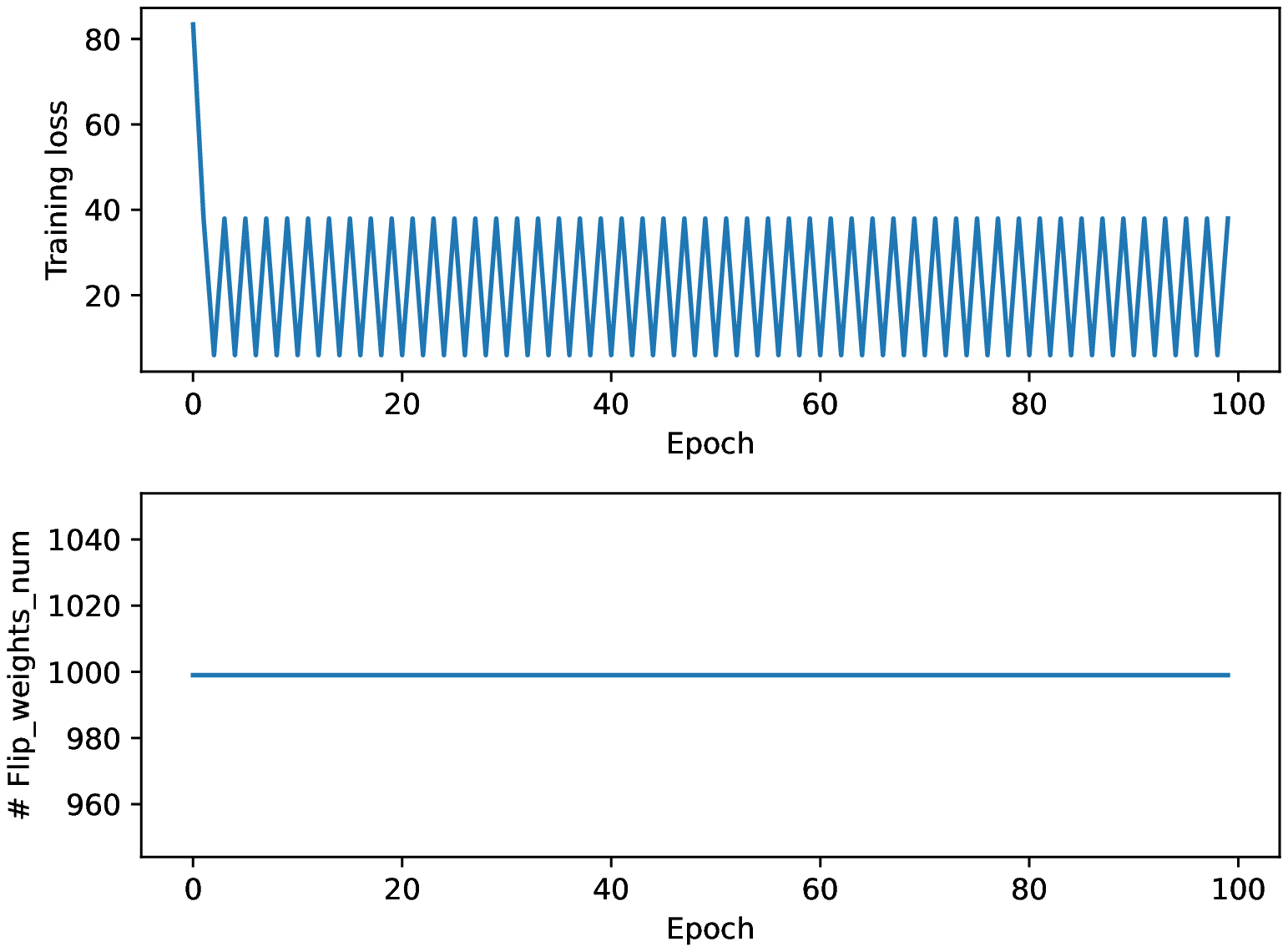}
            \caption{MAX and MSD}\label{fig: intro2_MAX MSD}
        \end{subfigure}
        \begin{subfigure}{0.45\linewidth}
            \includegraphics[width=1\linewidth]{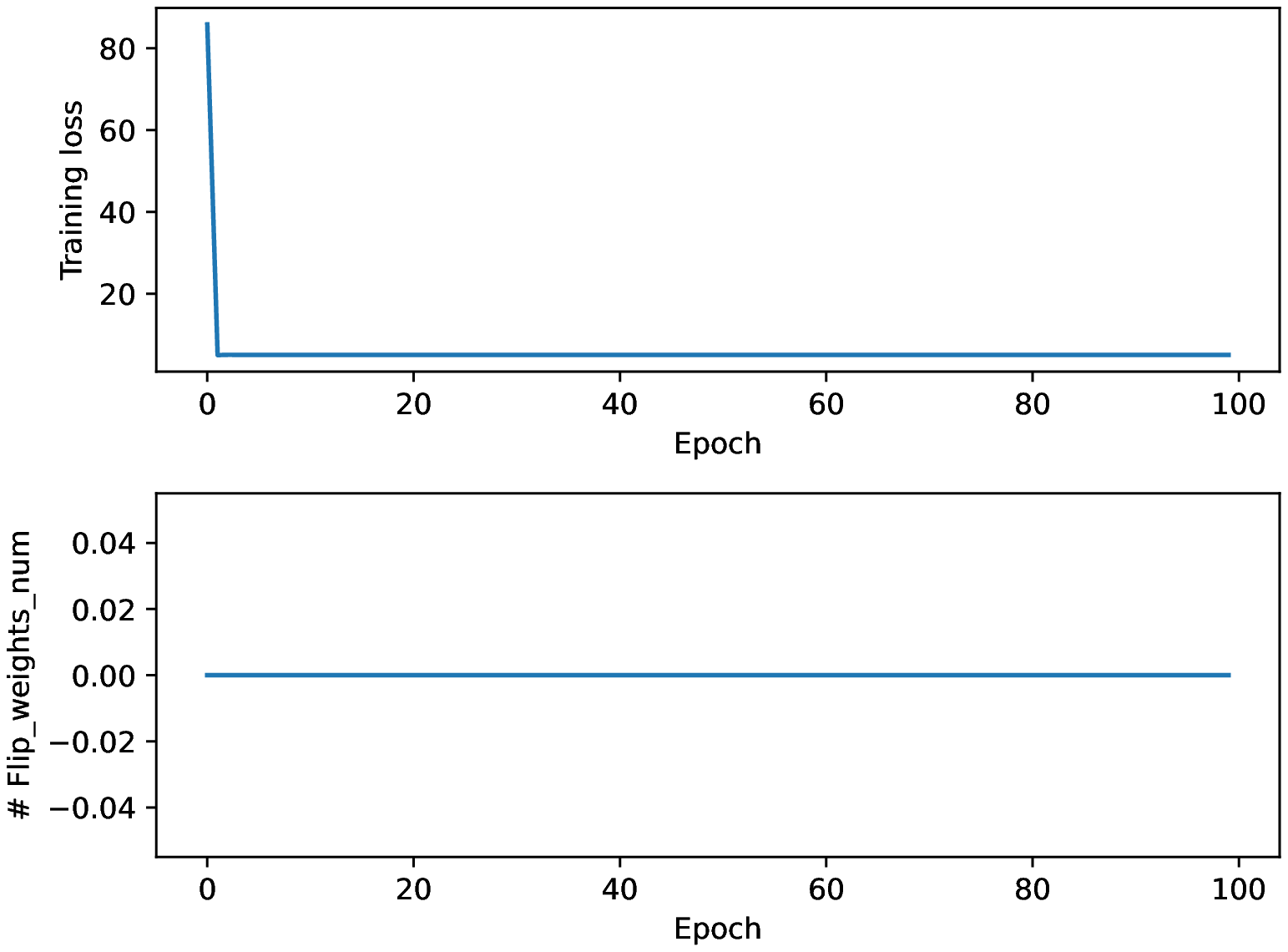}
            \caption{AVG}\label{fig: intro2_AVG}
        \end{subfigure}
        
        \caption{We illustrate the training loss and the number of weights that flip between two epochs. We set $\epsilon_1=4\mu, \epsilon_2=\epsilon_\infty=2\mu$. \Cref{fig: intro2_MAX MSD} shows the data of model trained with MAX and MSD using the SVM model (Sec~\ref{Sec: Data}) while \Cref{fig: intro2_AVG} shows the number of model trained with AVG.}
        \label{fig:intro2}
    \end{center}
\end{figure*}

\begin{figure*}[h]
    \begin{center}
        \begin{subfigure}{0.45\linewidth}
            \includegraphics[width=1\linewidth]{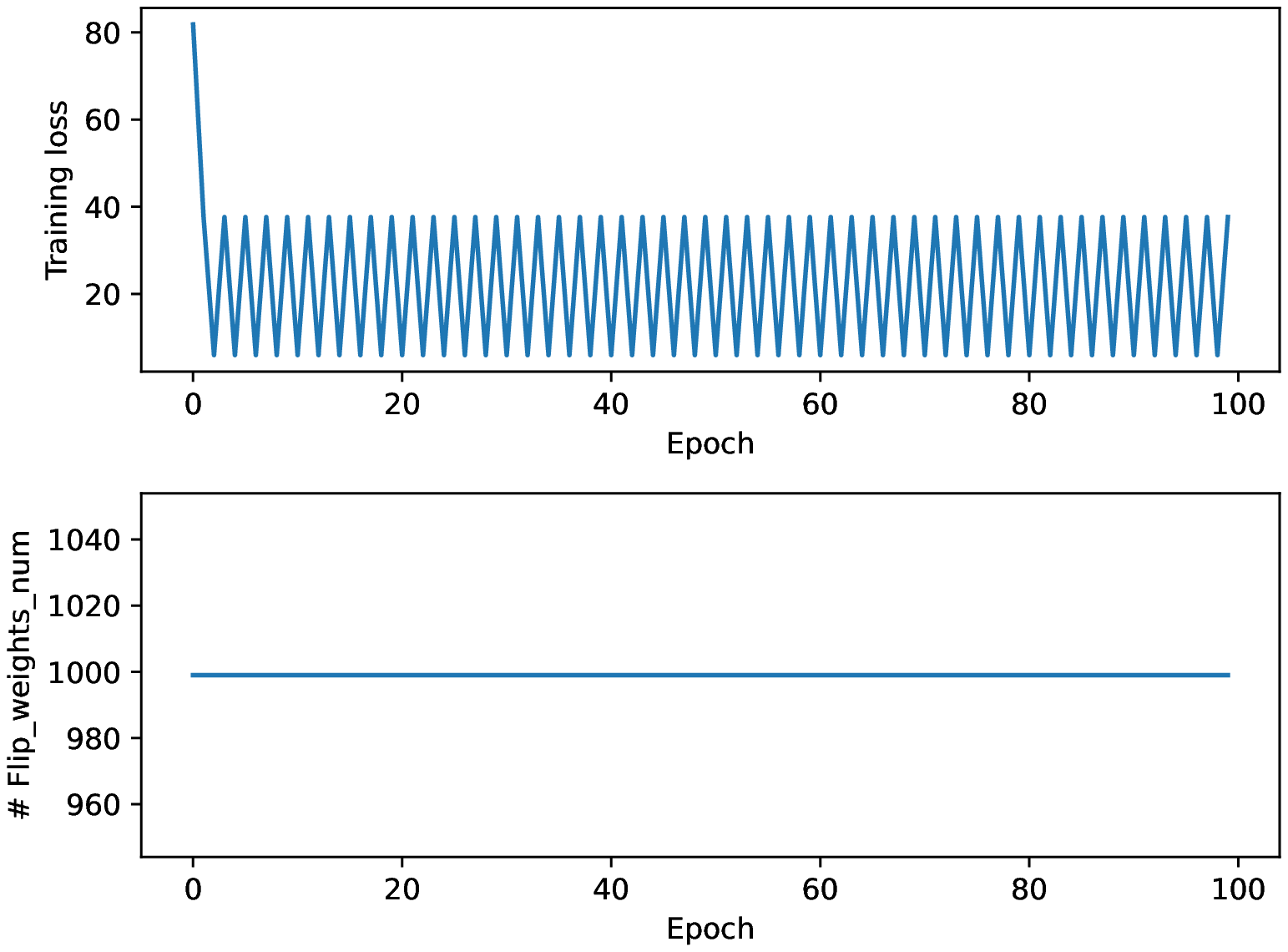}
            \caption{MAX and MSD}\label{fig: intro3_MAX MSD}
        \end{subfigure}
        \begin{subfigure}{0.45\linewidth}
            \includegraphics[width=1\linewidth]{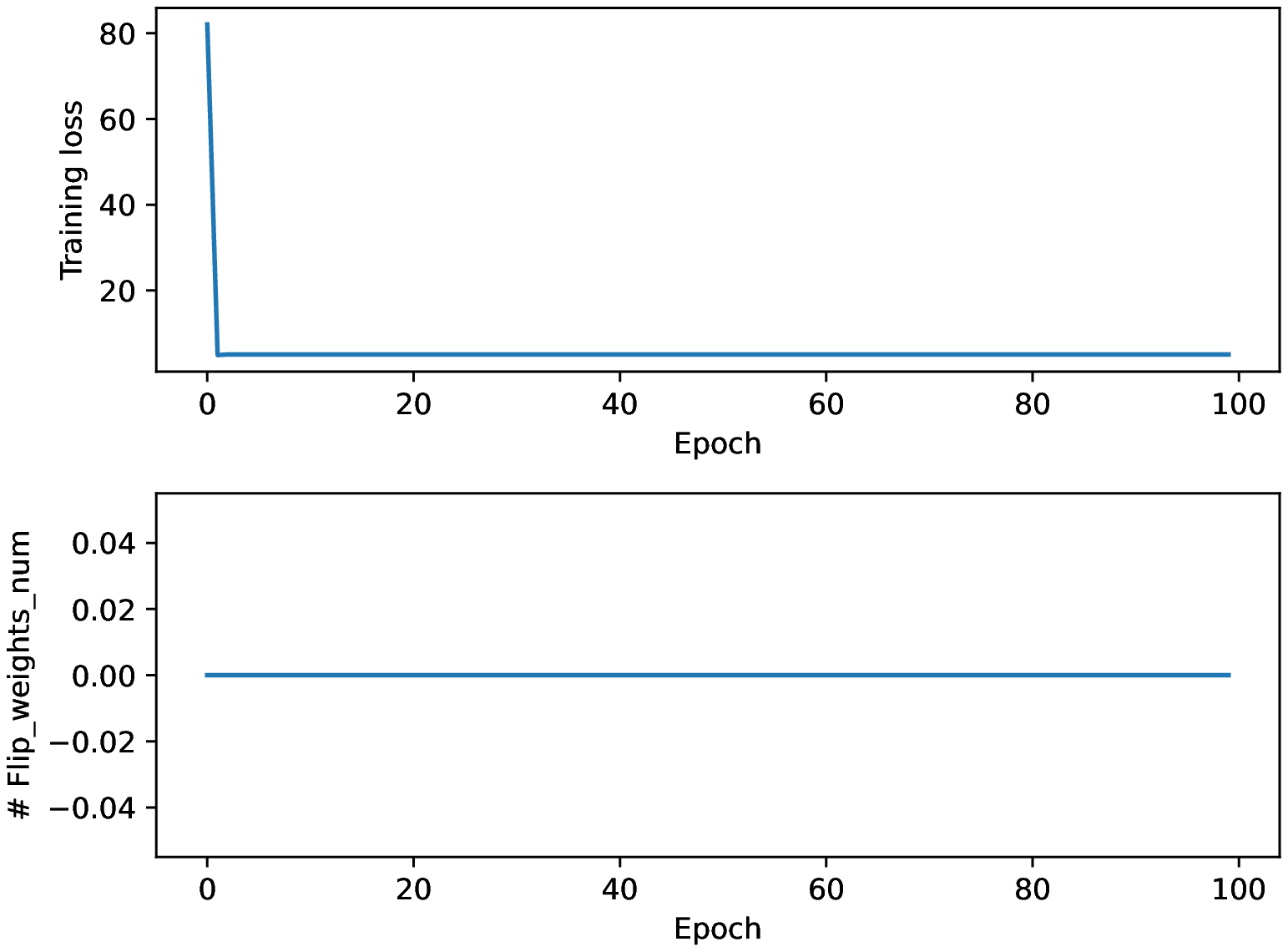}
            \caption{AVG}\label{fig: intro3_AVG}
        \end{subfigure}
        
        \caption{We illustrate the training loss and the number of weights that flip between two epochs. We set $\epsilon_2=4\mu, \epsilon_1=\epsilon_\infty=2\mu$. \Cref{fig: intro3_MAX MSD} shows the data of model trained with MAX and MSD using the SVM model (Sec~\ref{Sec: Data}) while
        \Cref{fig: intro3_AVG} shows the number of model trained with AVG.}
        \label{fig:intro3}
    \end{center}
\end{figure*}

Besides, we also conduct experiments verifying the conjecture that when $\ell_1$ or $\ell_2$ player dominates the bargaining game, the training procedure does not converge as well. For the case when $\ell_1$ dominates, we set $\epsilon_1=4\mu, \epsilon_2=\epsilon_\infty=2\mu$, while when $\ell_2$ dominates, we set $\epsilon_2=4\mu, \epsilon_1=\epsilon_\infty=2\mu$. We observe exactly the same curves as Figure~\ref{fig:intro1}, showing that when $\ell_1$ and $\ell_2$ dominates, the training procedure with MAX and MSD cannot converge while with AVG, training procedure does converge. We present the results in Figures~\ref{fig:intro2} and \ref{fig:intro3}.

\comment{
\subsection{Results on MNIST}

Table~\ref{tab:mnist-aa} shows the experimental results on MNIST with the AutoAttack repository. For the MNIST dataset, we require that the adapted epsilon are bigger than the half of and smaller that twice of the original epsilon to ensure that the update direction will not be influenced by an outlier who may have extremely unbalance gradients.

\begin{table}[h]
\centering
\caption{Summary of adversarial accuracy results for MNIST (higher is better). ``w. adaptive budget'' refers to employ our proposed method which enables adapting epsilon to avoid the player domination phenomenon while $\ell_1$ and $\ell_2$ means using $\ell_1$ and $\ell_2$ metrics to calculate the norm of gradients. ``AA'' refers to AutoAttack.}
\label{tab:mnist-aa}
\resizebox{\textwidth}{!}{%
\begin{tabular}{l|c|c|c|ccc|ccc|ccc}
\toprule
Models                  & $\ell_1$ & $\ell_2$ & \multicolumn{1}{c|}{$\ell_\infty$} & \multicolumn{3}{c|}{MAX}                                 & \multicolumn{3}{c|}{MSD}                                     & \multicolumn{3}{c}{AVG}                                    \\
w. adaptive budget     &          &          &                                    &      & $\ell_1$ (Ours)                & $\ell_2$ (Ours)                &      & $\ell_1$ (Ours)                  & $\ell_2$ (Ours)                  &      & $\ell_1$ (Ours)                & $\ell_2$ (Ours)                  \\\midrule\midrule
Clean Accuracy ($\%$)         & 97.2     & 99.1     & 99.2                               & 98.6 & 98.9                    & 98.9                    & 98.2 & 98.3                      & 98.9                      & 99.1 & 99.1                    & 99.1                      \\\midrule
$\ell_1$ AA Attack Robust Acc ($\%$)   & 42.6     & 46.9     & 2.6                                & 34.3 & 18.4$\downarrow$        & 16.9$\downarrow$        & 47.3 & 31.3$\downarrow$          & 28.6$\downarrow$          & 12.8 & \textbf{19.3$\uparrow$} & \textbf{18.4$\uparrow$}   \\\midrule
$\ell_2$ AA Attack Robust Acc ($\%$)     & 22.2     & 64.9     & 6.5                                & 64.6 & 59.2$\downarrow$        & 60.0$\downarrow$        & 65.3 & 62.6$\downarrow$          & 60.6$\downarrow$          & 63.3 & 56.9$\downarrow$        & 56.8$\downarrow$ \\\midrule
$\ell_\infty$ AA Attack Robust Acc ($\%$) & 0        & 0.1      & 8.8                                & 10.3 & \textbf{58.4$\uparrow$} & \textbf{59.2$\uparrow$} & 60.0 & \textbf{68.3$\uparrow$}   & \textbf{67.4$\uparrow$}   & 44.7 & \textbf{58.1$\uparrow$} & \textbf{56.2$\uparrow$}   \\\midrule
AA Attack All           & 0        & 0.1      & 2.1                                & 10.1 & \textbf{17.9$\uparrow$} & \textbf{16.2$\uparrow$} & 45.5 & 31.2$\downarrow$ & 28.4$\downarrow$ & 12.0 & \textbf{18.2$\uparrow$} & \textbf{17.3$\uparrow$}  \\\bottomrule
\end{tabular}%
}
\end{table}
}

\subsection{Results on CIFAR10}

Similar, for the CIFAR10 dataset, we require that the adapted epsilon are bigger than the half of and smaller that twice of the original epsilon.

\begin{figure*}[h]
   \begin{center}
               \includegraphics[width=0.28\linewidth]{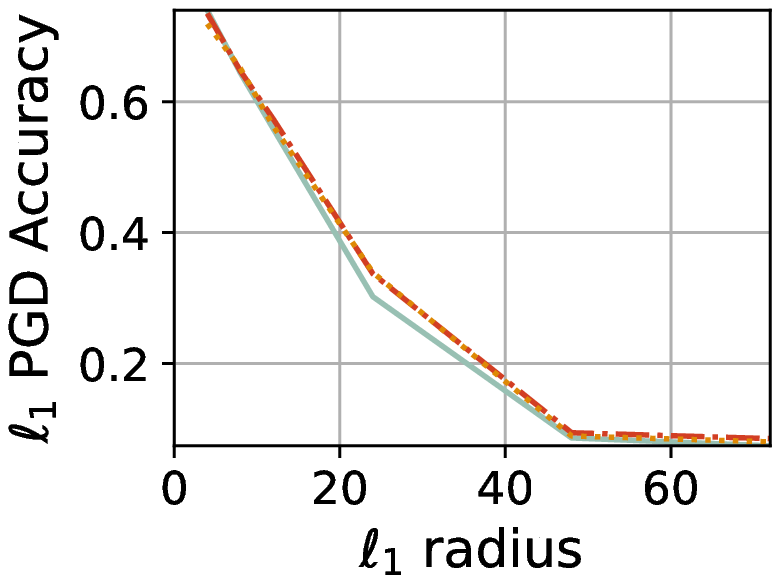}
      \includegraphics[width=0.28\linewidth]{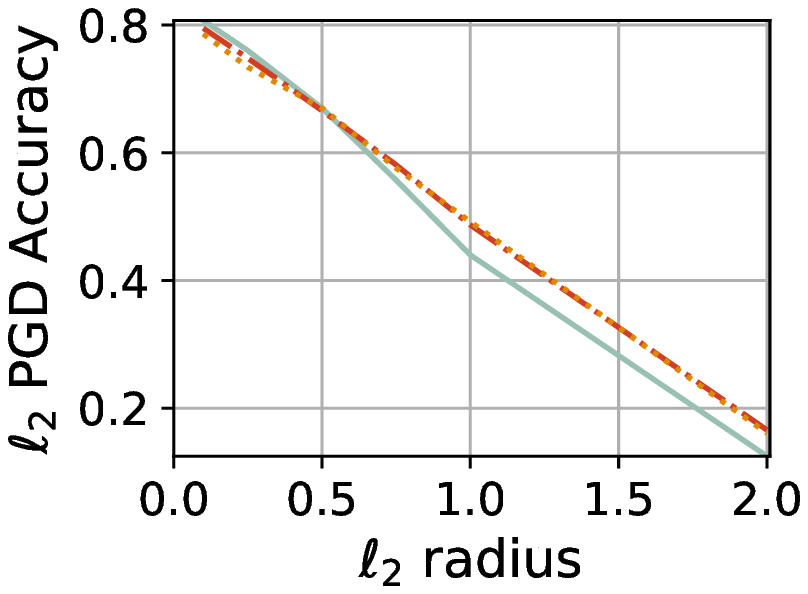}
      \includegraphics[width=0.28\linewidth]{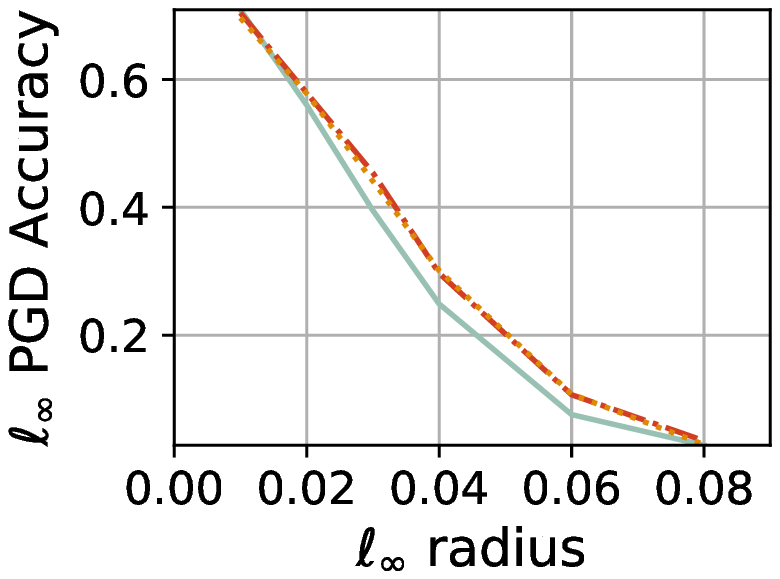}
      \includegraphics[width=0.1\linewidth]{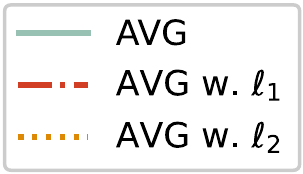}
      
      \includegraphics[width=0.28\linewidth]{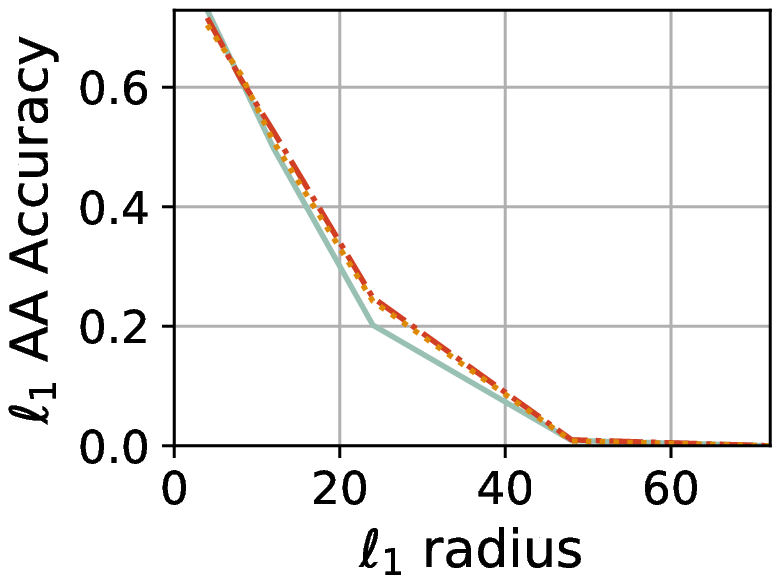}
      \includegraphics[width=0.28\linewidth]{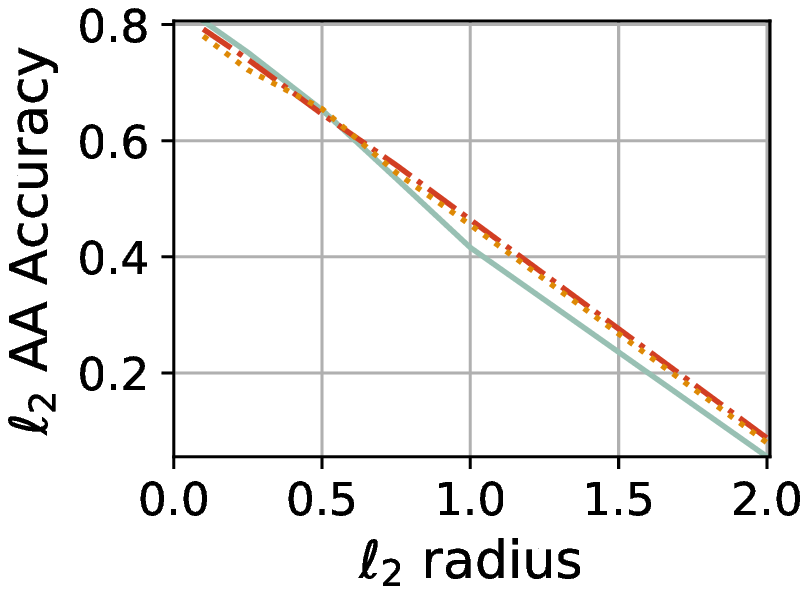}
      \includegraphics[width=0.28\linewidth]{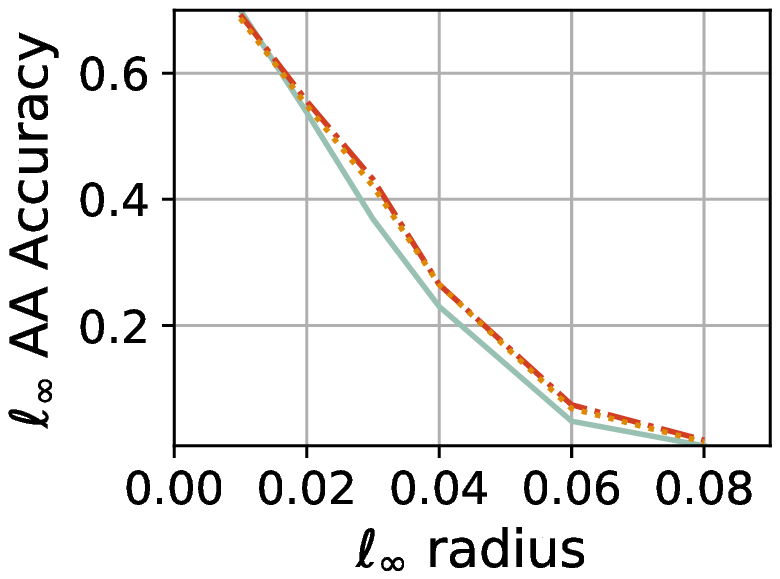}
      \includegraphics[width=0.1\linewidth]{pic/cifar/CIFAR_legend_AVG.pdf}

            \includegraphics[width=0.28\linewidth]{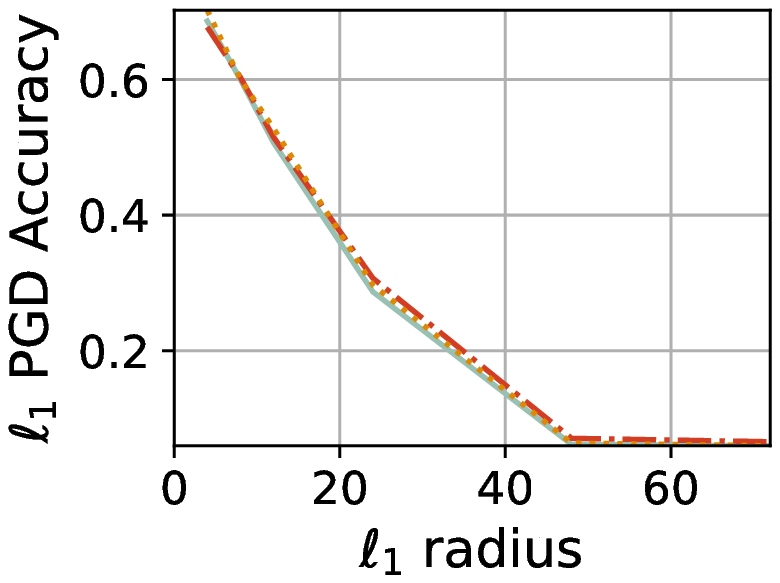}
      \includegraphics[width=0.28\linewidth]{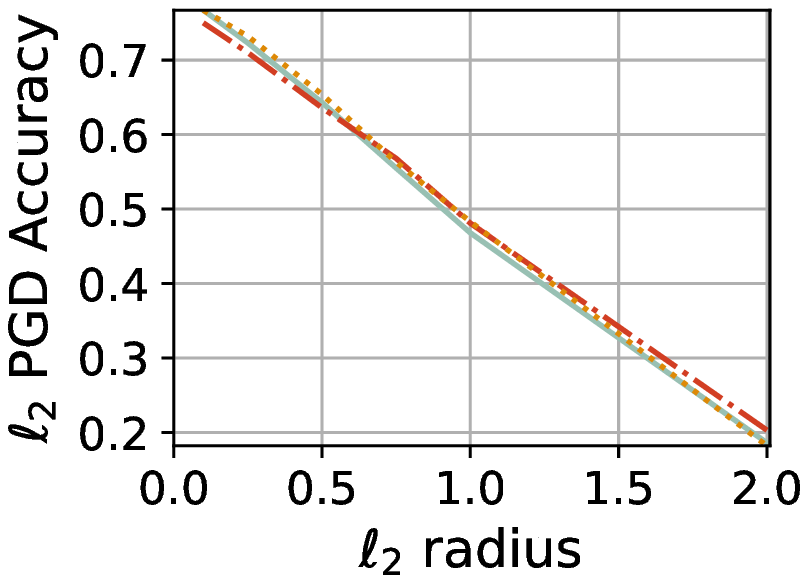}
      \includegraphics[width=0.28\linewidth]{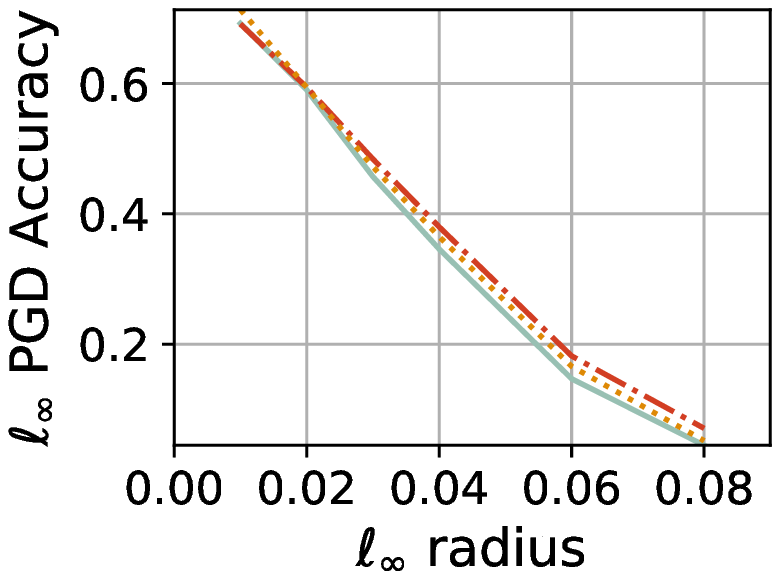}
      \includegraphics[width=0.1\linewidth]{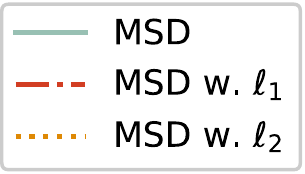}

      \includegraphics[width=0.28\linewidth]{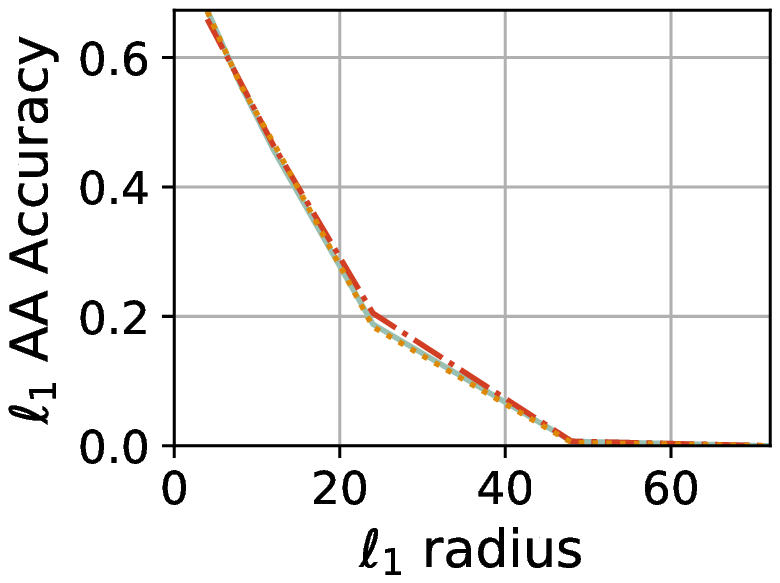}
      \includegraphics[width=0.28\linewidth]{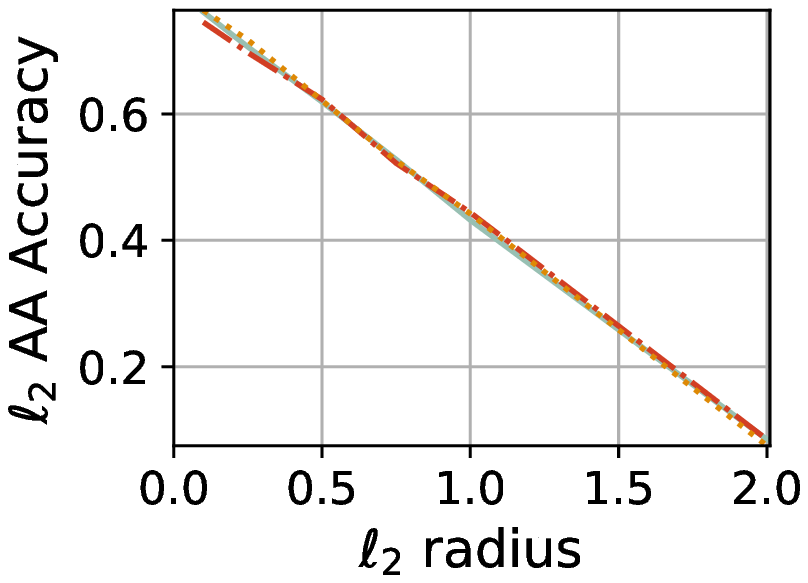}
      \includegraphics[width=0.28\linewidth]{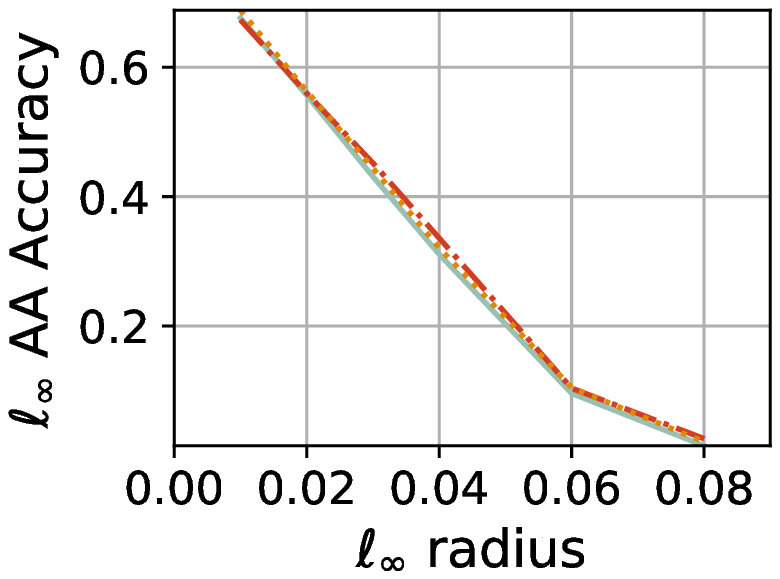}
      \includegraphics[width=0.1\linewidth]{pic/cifar/CIFAR_legend_MSD.pdf}
      
            \includegraphics[width=0.28\linewidth]{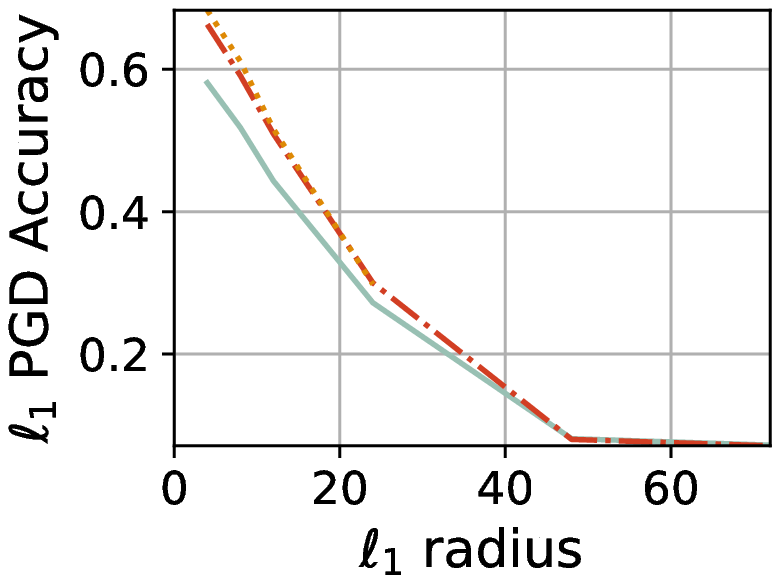}
      \includegraphics[width=0.28\linewidth]{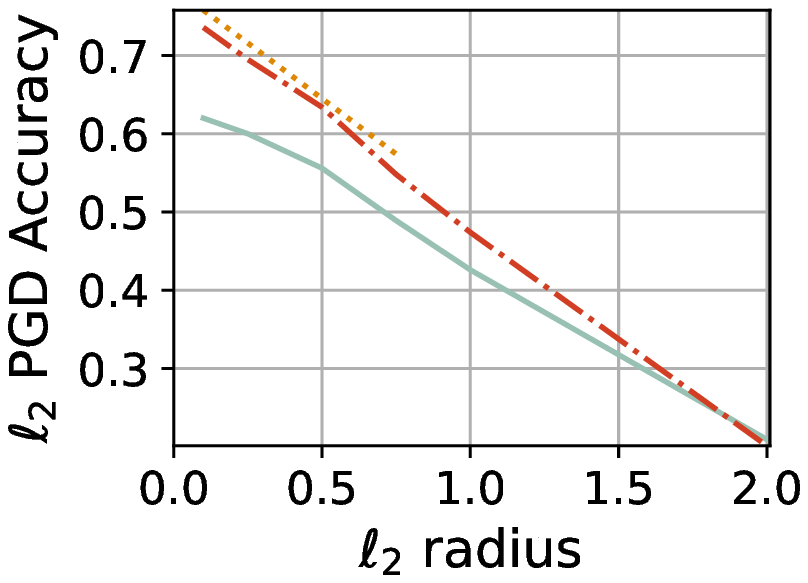}
      \includegraphics[width=0.28\linewidth]{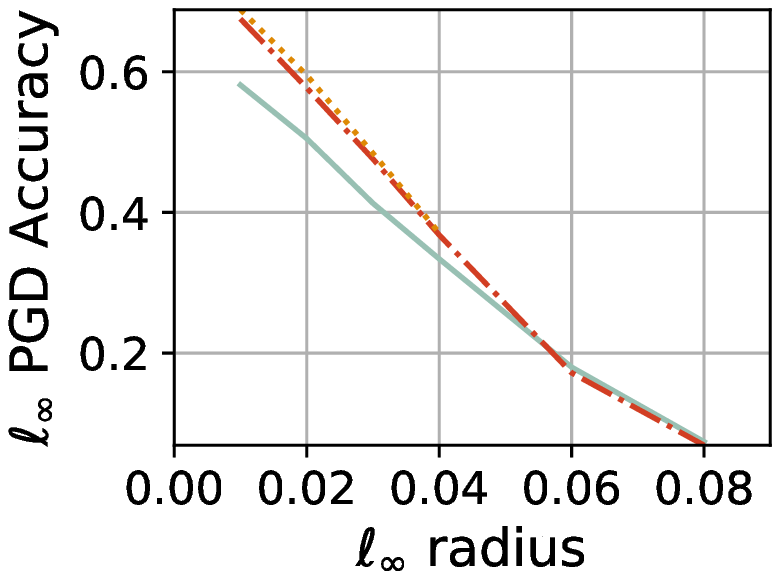}
      \includegraphics[width=0.1\linewidth]{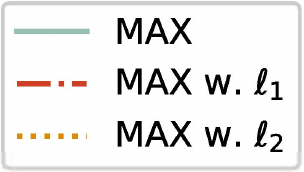}
      
        \includegraphics[width=0.28\linewidth]{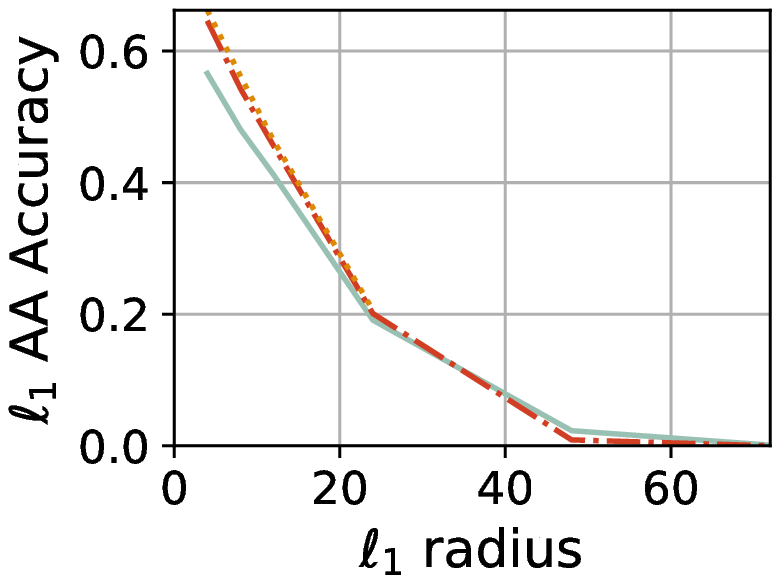}
      \includegraphics[width=0.28\linewidth]{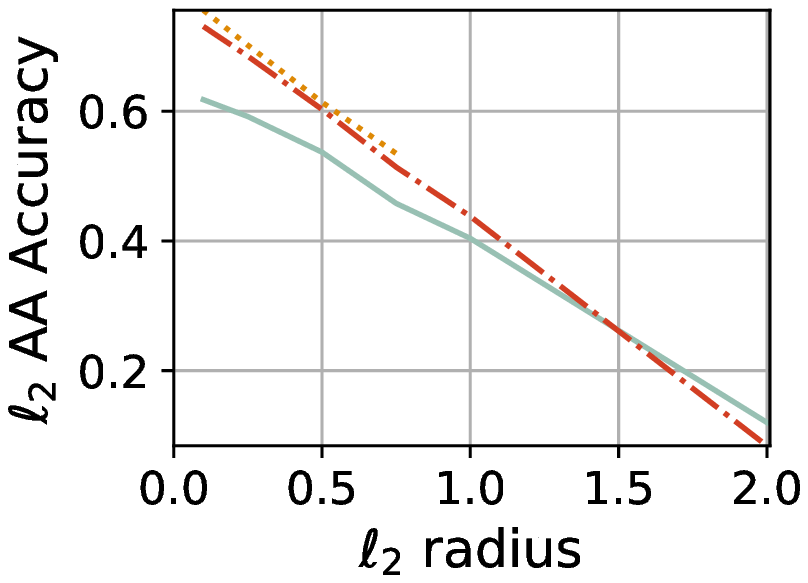}
      \includegraphics[width=0.28\linewidth]{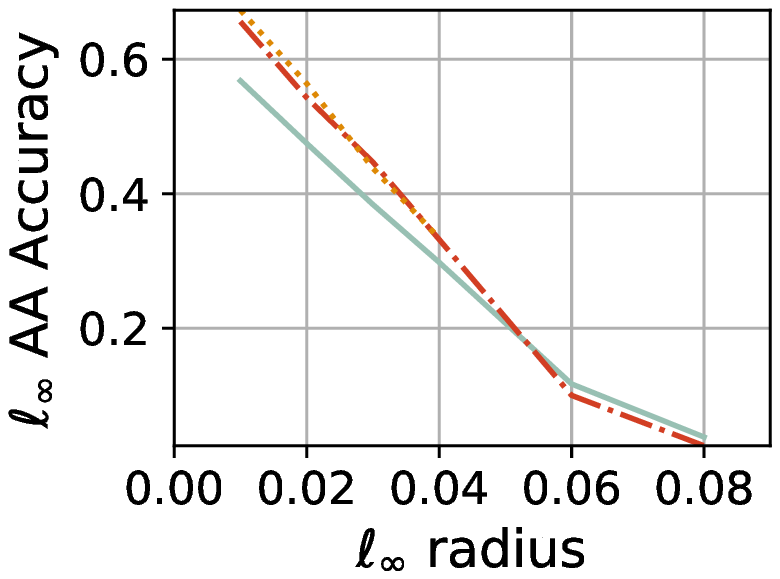}
      \includegraphics[width=0.1\linewidth]{pic/cifar/CIFAR_legend_MAX.pdf}

      \caption{
      Robustness curves show the adversarial accuracy on CIFAR-10 trained with MSD, AVG, and MAX against $\ell_1$ (left), $\ell_2$ (middle), and $\ell_\infty$ (right) PGD and AutoAttack (``AA'' in the figures) adversaries over a range of epsilon. 
      ``w. $\ell_1$'' and ``w. $\ell_2$'' are methods with AdaptiveBudget using $\ell_1$ or $\ell_2$ norms.
      }
      \label{fig: robustness curves cifar aa}
   \end{center}
\end{figure*}

\textbf{Robustness curves.} The robustness curves are shown in Figures~\ref{fig: robustness curves cifar aa}. The lines of MAX with either $\ell_1$ or $\ell_2$ norm-based AdaptiveBudget are higher than the lines without AdaptiveBudget. 
The gap between lines with the adaptive budget method and lines without is biggest when the budget of the adversary is small. 

\textbf{Norm choice in adaptive budget. }We notice that the choice of norm in the adaptive budget barely influences the robust accuracy as shown in Table~\ref{tab:cifar-pgd}. On both three methods, \ie, MAX, MSD, and AVG, our proposed adaptive budget is able to improve the performance with both $\ell_1$ and $\ell_2$ norms, while the difference between $\ell_1$ and $\ell_2$ norm is only $0.6\%$ and $1.7\%$, $0.4\%$ and $2.8\%$, $1.3\%$ and $0.9\%$ on MAX, MSD, and AVG against PGD and AutoAttack adversaries.

\subsection{Verification of avoiding player domination}
We record the maximum number of consecutive steps during which the gradient of a player is consistently higher than others on MNIST with deep neural networks. 
Quantitative results in Table~\ref{tab: player domination consecutive steps} indicate that, with \textit{AdaptiveBudget} on MNIST, no player is able to control the updates for more than \textcolor{blue}{1207} steps, whereas, with MSD, the $\ell_\infty$-player controlled the updates for \textcolor{blue}{6891} consecutive steps, respectively. 
Besides, the average number during which a player dominated is significantly decreased with \textit{AdaptiveBudget}. 
This demonstrates the effectiveness of \textit{AdaptiveBudget} in avoiding player domination.

\begin{table}[]
\centering
\caption{The robust accuracy of TRADES on MNIST and CIFAR-10.}
\label{tab: TRADES}
\begin{tabular}{l|cc}
\hline
TRADES           & MNIST & CIFAR-10 \\ \hline
$\ell_\infty$ AA      & 90.9  & 55.1    \\
$\ell_1$ AA      & 4.6   & 6.2     \\
$\ell_2$ AA & 11.2  & 60.5    \\ \hline
All AA           & 3.7   & 6.2     \\ \hline
\end{tabular}%
\end{table}

\begin{table*}[h!]
\centering
\caption{
This table displays the maximum number of consecutive steps (updates) during which the gradient of a single player is bigger than others on MNIST. 
``w. $\ell_1$'' and ``w. $\ell_2$'' refer to the algorithm with \textit{AdaptiveBudget}. 
``AVG'' refers to the average length of consecutive steps that player domination lasts. 
For example, if the training process consists of 5 steps and the dominant players are $[\ell_1, \ell_1, \ell_2, \ell_\infty, \ell_\infty]$, the longest consecutive steps for $\ell_1, \ell_2,$ and $\ell_\infty$ are 2, 1, and 2, and the average step (AVG) is 1.67.
}
\label{tab: player domination consecutive steps}
\begin{tabular}{l|ccc|ccc|ccc}
\toprule
                            & \multicolumn{3}{c|}{MSD}                                                                 & \multicolumn{3}{c|}{MAX}                                                                 & \multicolumn{3}{c}{AVG}                                                                \\
                            &       & w. $\ell_1$                           & w. $\ell_2$                             &      & w. $\ell_1$                             & w. $\ell_2$                            &      & w. $\ell_1$                             & w. $\ell_2$                           \\
                            \midrule
$\ell_1$-PGD adversary      & 60    & 61     \textcolor{red}{$\uparrow$}    & 27     \textcolor{blue}{$\downarrow$}   & 15   & 5     \textcolor{blue}{$\downarrow$}    & 9   \textcolor{blue}{$\downarrow$}     & 8    & 10      \textcolor{red}{$\uparrow$}     & 4      \textcolor{blue}{$\downarrow$} \\
$\ell_2$-PGD adversary      & 113   & 140     \textcolor{red}{$\uparrow$}   & 209      \textcolor{red}{$\uparrow$}    & 337  & 323      \textcolor{blue}{$\downarrow$} & 303     \textcolor{blue}{$\downarrow$} & 291  & 154      \textcolor{blue}{$\downarrow$} & 39   \textcolor{blue}{$\downarrow$}   \\
$\ell_\infty$-PGD adversary & 6891  & 1207   \textcolor{blue}{$\downarrow$} & 572     \textcolor{blue}{$\downarrow$}  & 52   & 45      \textcolor{blue}{$\downarrow$}  & 26    \textcolor{blue}{$\downarrow$}   & 32   & 26   \textcolor{blue}{$\downarrow$}     & 123    \textcolor{red}{$\uparrow$}    \\
\midrule
AVG                         & 18.77 & 10.44 \textcolor{blue}{$\downarrow$}  & 8.97     \textcolor{blue}{$\downarrow$} & 6.23 & 2.41 \textcolor{blue}{$\downarrow$}     & 2.41   \textcolor{blue}{$\downarrow$}  & 4.24 & 2.45   \textcolor{blue}{$\downarrow$}   & 2.33  \textcolor{blue}{$\downarrow$} \\
\bottomrule
\end{tabular}%
\end{table*}

\subsection{Are single-target robustness algorithms able to achieve multi-target robustness?}
In \Cref{tab:mnist-pgd,tab:cifar-pgd}, we illustrate that three typical single-target methods, \ie, $\ell_1$, $\ell_2$, and $\ell_\infty$, are not able to maintain multi-target robustness as the overall robust accuracy drops a lot. 
To compare with more methods, 
we conduct experiments on a representative single-target method, \ie, TRADES~\cite{zhangTheoreticallyPrincipledTradeoff2019}. 
We found that TRADES fails to defend $\ell_1$, $\ell_2$, and $\ell_\infty$ AutoAttack's adversaries simultaneously. 
As shown in \Cref{tab: TRADES}, the overall accuracies of \href{https://github.com/yaodongyu/TRADES}{TRADES} on MNIST and CIFAR-10 are only 3.9\% and 6.2\%, respectively.

\end{document}